\documentclass[10pt,twocolumn,letterpaper]{article}
\usepackage[pagenumbers]{cvpr}

\usepackage{tabularx}
\usepackage{multirow}
\usepackage{colortbl}
\usepackage{algorithm}
\usepackage{algpseudocode}
\usepackage{placeins}
\definecolor{arxivlinkblue}{rgb}{0.21,0.49,0.74}
\usepackage[colorlinks,allcolors=arxivlinkblue]{hyperref}

\newcommand{\method}{RANGER}

\title{RANGER: Learning Socially Aware Visual Navigation Skills via Guided Exploration and Refinement}
\author{%
  \begin{tabular}{@{}c@{\qquad}c@{\qquad}c@{\qquad}c@{}}
    Yuhan Pang\textsuperscript{1} & Bingyi Xia\textsuperscript{1}
    & Yizhong Chen\textsuperscript{1} & Zhewei Chen\textsuperscript{1}\\
    Han Bao\textsuperscript{1} & Fang Liu\textsuperscript{2}
    & Wenjun Xu\textsuperscript{3} & Jiankun Wang\textsuperscript{1}
  \end{tabular}\\[10pt]
  {\normalsize\textsuperscript{1}\,Southern University of Science and Technology}\\
  {\normalsize\textsuperscript{2}\,Shenzhen University
  \qquad\textsuperscript{3}\,Peng Cheng Laboratory}%
}

\makeatletter
\patchcmd{\@maketitle}{\vskip .375in}{\vskip .18in}{}%
  {\PackageError{arxiv}{Could not adjust title spacing}{}}
\makeatother

\begin{document}
\begingroup
\setlength{\dblfloatsep}{10pt plus 2pt minus 2pt}
\setlength{\dbltextfloatsep}{12pt plus 2pt minus 3pt}
\setlength{\textfloatsep}{12pt plus 2pt minus 3pt}
\maketitle

\newcommand{\arxivprojectpage}{%
  The project page is available at
  \href{https://yhpang02.github.io/ranger/}{\mbox{yhpang02.github.io/ranger/}}.}
\begin{abstract}

Visual navigation in human environments requires robots to acquire diverse
social interaction skills beyond geometric collision avoidance.
Although generative navigation policies can represent multimodal trajectory
distributions, learning such behaviors remains challenging due to the scarcity
of human--robot interaction demonstrations.
Reinforcement post-training can further refine these policies, but often
struggles to explore desirable behaviors beyond the pretrained distribution.
To address these challenges, we propose \method{}, a unified framework
for learning diverse social navigation skills through interaction-rich
pretraining and guided reinforcement post-training.
We first collect interaction-rich demonstrations using multiple navigation
experts and articulated pedestrians to fine-tune a pedestrian-aware visual
navigation policy.
Building on this policy, we introduce Ranger-GRPO, which uses differentiable
skill-specific energies to guide stochastic flow sampling toward promising
interaction behaviors during online post-training.
Trajectory rewards then provide group-relative advantages to reinforce
and distill these behaviors into the policy.
Experiments on dynamic navigation benchmarks show that \method{}
improves navigation effectiveness and interaction safety over baselines,
while a single policy exhibits multiple social navigation skills
and transfers directly to real-robot navigation in human environments.
\ifdefined\arxivprojectpage\ \arxivprojectpage\fi

\end{abstract}

\section{Introduction}

\begin{figure}[t]
    \centering
    \includegraphics[width=\columnwidth]{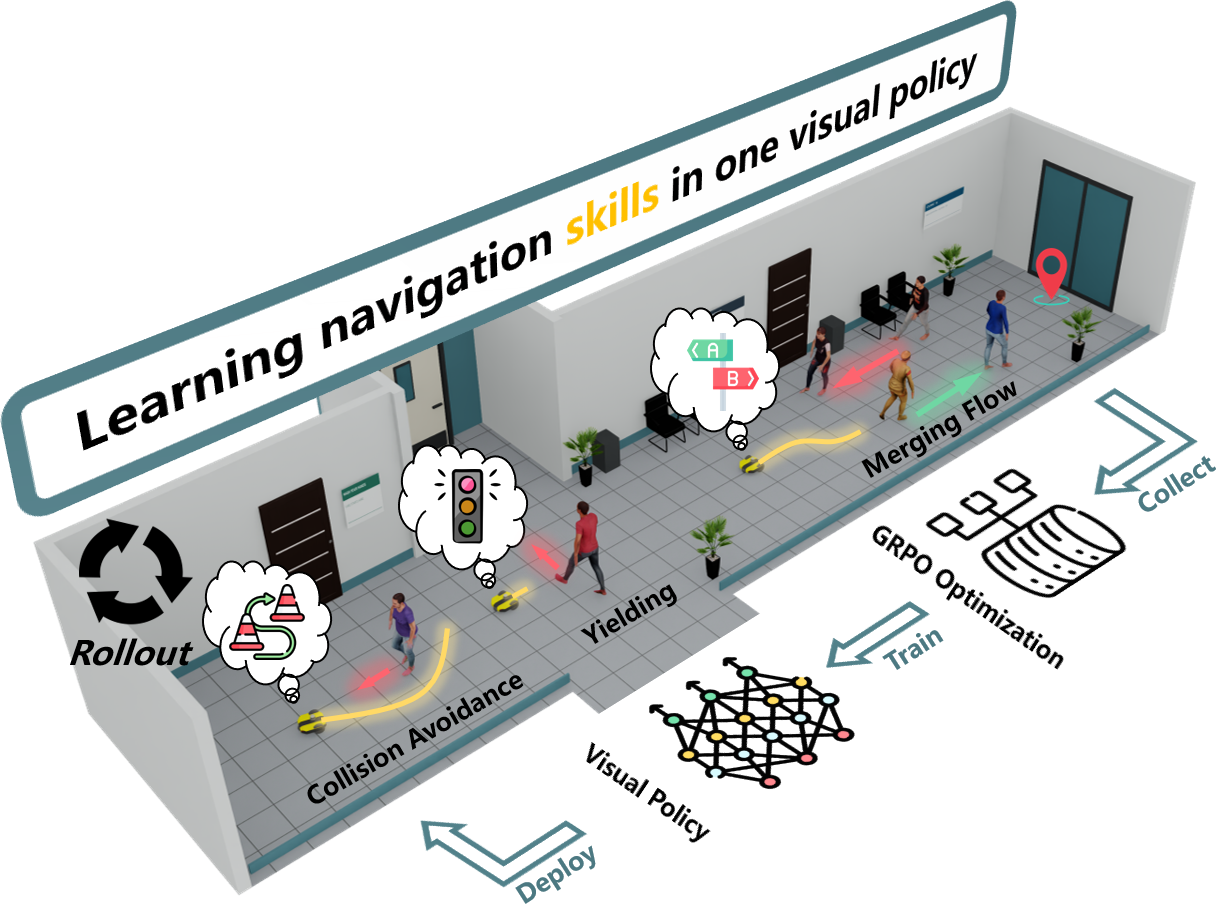}
    \caption{RANGER learns multiple social navigation skills within a unified visual policy. Closed-loop rollouts capture diverse human–robot interactions, including collision avoidance, yielding, and merging into pedestrian flows. The resulting trajectories are used for Ranger-GRPO post-training.}
    \label{fig:background}
\end{figure}

Visual navigation enables generalizable robot autonomy by
learning goal-conditioned policies directly from onboard
observations, reducing reliance on hand-crafted rules.
Large-scale learning and generative policy modeling have
shown potential for generalization across environments and
robot embodiments~\cite{shah2023gnm,shah2023vint,
sridhar2024nomad,cai2026navdp,gong2026flux}.
However, navigating among people requires more than geometric
obstacle avoidance. Robots need diverse social navigation
skills, including collision avoidance, yielding, merging with
pedestrian flows, and passing by groups, to move safely and
efficiently while minimizing disruption to human activities.

Learning a pedestrian-aware navigation policy requires interaction-rich demonstrations that capture diverse behavioral responses and temporal dependencies between robot and pedestrian actions~\cite{karnan2022scand,hirose2024sacson, brayan2026navwareset}, rather than collision-free paths derived solely from geometric occupancy. However, large-scale interaction-rich demonstrations remain scarce, limiting the behavioral coverage of imitation-based pretraining~\cite{sridhar2024nomad, cai2026navdp}. Reinforcement-learning post-training can further acquire and refine social navigation skills through task-specific rewards~\cite{gong2026flux, huang2026grace}, but its effectiveness is constrained when the pretrained policy rarely samples desirable interaction behaviors. Therefore, the key challenge is to efficiently explore these underrepresented behaviors and consolidate them into a unified policy capable of exhibiting multiple social navigation skills.

To address these challenges, we introduce \textbf{\method{}} (\textbf{R}einforcement-\textbf{A}ugmented \textbf{N}avigation via \textbf{G}uided \textbf{E}xploration and \textbf{R}efinement), a generative framework for learning diverse social navigation skills within a single visual policy, as illustrated in Fig.~\ref{fig:background}. We first construct an interaction-rich data generation pipeline with articulated pedestrians to provide temporally continuous human--robot interactions and diverse demonstrations for pretraining a pedestrian-aware navigation policy. Building on this policy, we propose Ranger-GRPO to acquire and refine social navigation skills through guided reinforcement post-training. Differentiable skill energies guide stochastic flow sampling toward promising interaction behaviors that are unlikely under the pretrained policy, while trajectory rewards provide group-relative advantages to reinforce and distill these behaviors into the unguided flow policy. Through this guided refinement, the pretrained policy acquires diverse, context-appropriate social navigation behaviors without requiring explicit skill labels, privileged pedestrian information, or guidance during inference.

Experiments on dynamic navigation benchmarks demonstrate that \method{} improves navigation effectiveness and interaction safety over existing baselines. Furthermore, a single \method{} policy exhibits multiple social navigation skills and transfers directly to real-world human--robot interactions without additional fine-tuning.
Our contributions are threefold:

\begin{itemize}
    \item We develop an interaction-rich data generation pipeline for
    dynamic visual navigation with articulated pedestrians, preserving both
    the geometry and timing of demonstrated interactions.
    
    \item We introduce Ranger-GRPO, which uses gradients of differentiable
    skill energies to guide reverse-time SDE sampling toward favorable
    interaction behaviors, and uses trajectory rewards in
    group-relative policy optimization to distill the resulting guided
    trajectories into the unguided flow policy.
    
    \item We realize multi-skill social navigation within a single visual
    policy and validate its generalization through closed-loop evaluation and
    real-robot deployment across diverse environments.
\end{itemize}

\section{Related Work}

\subsection{Visual Navigation}
Learning-based visual navigation has progressively moved from
geometry-centric navigation in largely static scenes to dynamic
and socially aware interaction. Early methods such as
iPlanner~\cite{yang2023iplanner} and
ViPlanner~\cite{roth2024viplanner} learn local navigation
from geometric and semantic supervision, while generalist
models such as GNM~\cite{shah2023gnm} and
ViNT~\cite{shah2023vint} improve transfer across
environments and robot embodiments through heterogeneous
navigation data. Recent approaches further scale visual
navigation learning using web videos
~\cite{liu2025citywalker,xia2026wildnav}.
Trajectory generation methods, including
NoMaD~\cite{sridhar2024nomad},
NavDP~\cite{cai2026navdp},
NaviBridger~\cite{ren2025navibridger},
FlowNav~\cite{gode2025flownav},
GeniNav~\cite{chen2026geninav},
S2E~\cite{he2026s2e}, and
STRNet~\cite{ren2026strnet}, further improve
navigation through generative modeling and
spatiotemporal visual representations.
More recent studies extend visual navigation to dynamic
human environments, emphasizing pedestrian-aware
motion planning and socially compliant interactions
~\cite{gong2026flux,hirose2024sacson,
song2025vlmsocialnav,bao2026icrowdnav,chen2026socialnav}.
Despite these advances, learning diverse social navigation
skills remains constrained by the availability of
interaction-rich demonstrations
~\cite{karnan2022scand,brayan2026navwareset}.
Collecting temporally continuous demonstrations that
capture diverse pedestrian motions and long-horizon
interactions, such as yielding, waiting, passing, and
recovery, is costly and difficult to scale.
Consequently, supervised generative policies may cover
common collision-free motion while underrepresenting
the interaction behaviors required in complex human
environments.

\subsection{Post-Training Generative Policies}
Although imitation learning provides an effective initialization for
generative policies, it primarily reproduces demonstrated behaviors and
offers limited ability to acquire interaction skills beyond the
demonstration distribution. Existing reward-driven
post-training methods broadly follow two paradigms. The first incorporates
reward feedback into generative learning objectives, assigning higher
weights to promising trajectories through reward- or value-weighted
regression and matching~\cite{yang2026xnavdp,zhang2026sidp,zheng2026unleash,huang2026grace}.
The second treats iterative generation as a stochastic policy and performs
policy-gradient optimization using likelihood ratios between current and
behavior policies at individual generation
steps~\cite{ren2025dppo,liu2025flowgrpo,gong2026flux,sheng2026beyond,chen2026socialnav,zou2025diffusiondrivev2}.
Despite these advances, both paradigms rely heavily on the quality and
diversity of sampled trajectories. Reward-weighted learning may
reinforce behaviors already dominant in the pretrained policy, while
likelihood-ratio optimization can be limited when exploration fails to
reach behaviorally distinct solutions. 
Recent studies have explored multimodal and adaptive
exploration~\cite{li2026plannerrft} and cost- or
gradient-guided sampling
~\cite{zeng2025navidiffusor,yan2025m2diffuser,
carvalho2025mpd,zheng2025diffusionplanner}
to generate more diverse and promising trajectories.
However, guidance-based methods typically rely on
inference-time guidance rather than transferring
guided behaviors into an unguided policy.


\begin{figure*}[t]
    \centering
    \includegraphics[width=\textwidth]{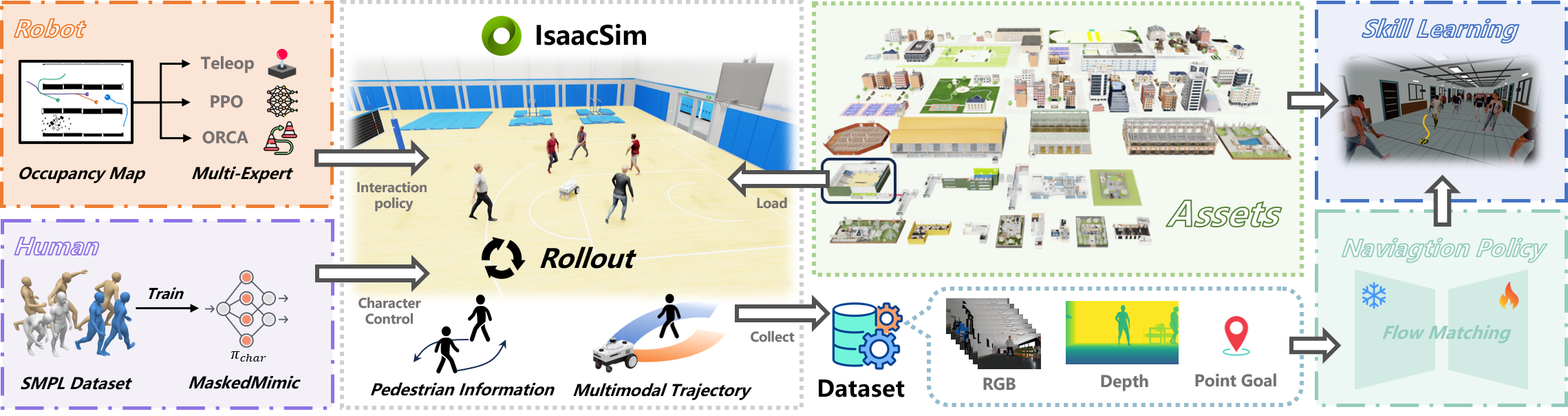}
    \caption{Interaction-rich data generation pipeline and policy training. Multiple navigation experts interact with articulated pedestrians in IsaacSim to collect synchronized multimodal navigation data and ground-truth pedestrian trajectories across diverse environments. The resulting dataset is then used to train the flow-based navigation policy and support subsequent social-skill learning.}
    \label{fig:overview}
    \vspace{-3mm}
\end{figure*}

\section{Method}
\label{sec:method}

\method{} is a generative framework for learning diverse social navigation
skills within a unified visual policy. The framework consists of an
interaction-rich data generation pipeline for skill-oriented pretraining
(Sec.~\ref{sec:data_generation}), a pedestrian-aware flow policy with a
trajectory critic (Sec.~\ref{sec:model}), and Ranger-GRPO for guided
reinforcement post-training and multi-skill refinement
(Sec.~\ref{sec:flow_grpo}).

\subsection{Interaction-Rich Data Generation}
\label{sec:data_generation}

Dynamic navigation requires demonstrations that capture collision avoidance and temporally coherent human-robot interactions across diverse pedestrian motion patterns.
Building on NavIsaacLab~\cite{xia2026navisaaclab}, we construct an
interaction-rich data generation pipeline that combines robot
demonstrations with articulated pedestrian interactions, as illustrated in
Fig.~\ref{fig:overview}.
In total, we collect
$94.05$\,km of navigation demonstrations from 50 indoor and outdoor
OmniGibson~\cite{li2022behavior} scenes.

\noindent\textbf{Articulated Pedestrian Simulation.}
To generate realistic dynamic human--robot interactions,
we simulate articulated pedestrians with physically plausible
and temporally coherent motion.
Each pedestrian $i$ follows a global route $\mathcal P_i$.
At each simulation step $k$, ORCA~\cite{vandenberg2011orca}
computes local collision-avoiding motion targets based on
the route and current simulator state.
These targets are converted into sparse conditions
$\mathbf g^{\mathrm{sparse}}_{i,k}$, comprising target positions
and orientations at upcoming timesteps and a longer-horizon
target.
A MaskedMimic controller~\cite{tessler2024maskedmimic}
generates full-body control actions conditioned on these
targets and the pedestrian's current body state:
\begin{equation}
\mathbf u^{\mathrm{char}}_{i,k}
\sim
\pi_{\mathrm{char}}\!\left(
\cdot\mid
\mathbf s^{\mathrm{body}}_{i,k},
\mathbf g^{\mathrm{sparse}}_{i,k}
\right),
\label{eq:pedestrian_generation}
\end{equation}
where $\mathbf s^{\mathrm{body}}_{i,k}$ denotes the current
pedestrian body state.

\noindent\textbf{Multi-Expert Demonstrations.}
Dynamic interaction is inherently multimodal: under the same visual
configuration, multiple collision-free behaviors such as passing,
yielding, waiting, or detouring may all be valid. To cover this
behavioral diversity, we combine demonstrations from a reactive ORCA
expert, a privileged PPO~\cite{schulman2017ppo} expert, and human teleoperation with a ratio
of $6{:}3{:}1$, respectively. Their complementary demonstrations are
stored in a unified RGB-D, point-goal, and trajectory format together
with training-only privileged geometry and future pedestrian tracks.

\subsection{Pedestrian-Aware Flow Policy}
\label{sec:model}
We train a pedestrian-aware flow policy through supervised
fine-tuning (SFT) on interaction-rich demonstrations.
The left half of Fig.~\ref{fig:ranger_framework}
illustrates its visual, pedestrian, and goal conditioning pathways.

\noindent\textbf{Observation and Action Representation.}
At navigation decision $k$, the observation
$\mathbf o_k=(I_{k-L+1:k},D_k,\mathbf g_k)$ contains $L$ historical
RGB frames, the current metric-depth image, and a point goal
$\mathbf g_k\in\mathbb R^2$.
The policy generates a normalized action sequence
$\mathbf a_k\in\mathbb R^{H\times3}$ from Gaussian noise conditioned
on $\mathbf o_k$. Each action contains planar displacement and
heading increments
$(\Delta x_{k,h},\Delta y_{k,h},\Delta\psi_{k,h})$.
A differentiable operator
$\boldsymbol\tau_k=\mathcal T(\mathbf a_k)\in\mathbb R^{H\times2}$
denormalizes and integrates the action sequence into $H$ future
waypoints. The goal, robot trajectory, and pedestrian positions are
expressed in the camera coordinate frame fixed at decision $k$.
A model-predictive controller tracks the generated trajectory, and
the policy replans as new observations arrive.

\noindent\textbf{Pedestrian-Aware Representation Learning.}
Let $\mathbf Z_k\in\mathbb R^{128\times384}$ denote the fused RGB-D
tokens and $\mathbf Q\in\mathbb R^{M\times384}$ denote a set of learnable
pedestrian queries. Pedestrian representations are extracted through
cross-attention:
\begin{equation}
\mathbf Z_k^{\mathrm{ped}}
=
\operatorname{LN}\!\left(
\mathbf Q+
\operatorname{MHA}(\mathbf Q,\mathbf Z_k,\mathbf Z_k)
\right),
\label{eq:pedestrian_tokens}
\end{equation}
where $\mathbf Z_k^{\mathrm{ped}}\in\mathbb R^{M\times384}$ contains
the instance-level pedestrian tokens.
Three lightweight prediction heads estimate each pedestrian's
current 2-D position, future displacement sequence, and existence
probability. Future positions are obtained by cumulatively
integrating the predicted displacements.
We use Hungarian assignment to match the unordered queries to valid
pedestrian trajectories. The matching cost combines current-position
and masked future-position errors, with assignments excluded from
gradient computation. Matched queries receive Smooth-$L_1$
supervision for position and motion prediction, while existence is
supervised using binary cross-entropy. The auxiliary loss is:
\begin{equation}
\mathcal L_{\mathrm{ped}}
=
\lambda_{\mathrm{cur}}\mathcal L_{\mathrm{cur}}
+
\lambda_{\mathrm{delta}}\mathcal L_{\mathrm{delta}}
+
\lambda_{\mathrm{exist}}\mathcal L_{\mathrm{exist}}.
\label{eq:pedestrian_loss}
\end{equation}

\begin{figure*}[t]
    \centering
    \includegraphics[width=\textwidth]{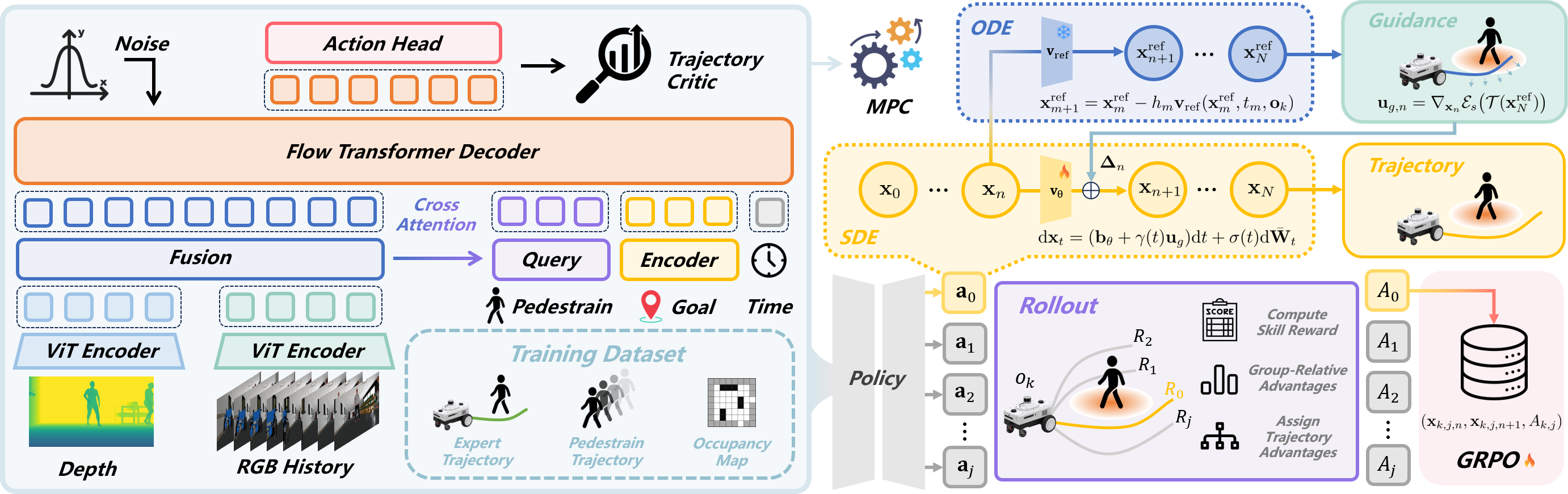}
    \caption{Overview of the RANGER architecture and training framework.
    \textbf{Left:} The pedestrian-aware flow policy integrates RGB-D features,
    learned pedestrian tokens, and goal conditioning to generate action
    trajectories from noise.
    \textbf{Right:} Ranger-GRPO uses skill-energy gradients
    from a frozen SFT reference model to guide stochastic SDE sampling,
    then evaluates sampled trajectories with skill-specific rewards
    and refines the unguided policy through group-relative
    policy optimization.}
    \vspace{-3mm}
    \label{fig:ranger_framework}
\end{figure*}

\noindent\textbf{Pedestrian-Conditioned Flow Matching.}
The pedestrian tokens are appended to the decoder conditioning memory
$[\mathbf z_t,\mathbf z_g,\mathbf Z_k,\mathbf Z_k^{\mathrm{ped}}]$,
where $\mathbf z_t$ and $\mathbf z_g$ denote the flow-time and goal
embeddings. Learned positional embeddings distinguish individual
pedestrian tokens. The flow decoder directly uses these latent
representations rather than the auxiliary pedestrian predictions.
Let $d=3H$ and denote the vectorized action by
$\mathbf a\in\mathbb R^d$.
Given $\boldsymbol\epsilon\sim\mathcal N(\mathbf0,\mathbf I_d)$ and
$t\sim\mathcal U(0,1)$, we use the interpolation
$\mathbf x_t=(1-t)\mathbf a+t\boldsymbol\epsilon$.
The conditional velocity field is trained by:
\begin{equation}
\mathcal L_{\mathrm{FM}}
=
\mathbb E_{\mathbf a,\boldsymbol\epsilon,t}
\left[
\left\|
\mathbf v_\theta(\mathbf x_t,t,\mathbf o)
-(\boldsymbol\epsilon-\mathbf a)
\right\|_2^2
\right].
\label{eq:flow_matching_loss}
\end{equation}
Generation integrates
$\mathrm d\mathbf x_t
=\mathbf v_\theta(\mathbf x_t,t,\mathbf o)\,\mathrm dt$
backward from $t=1$ (noise) to $t=0$ (actions).

\noindent\textbf{Trajectory Critic.}
For candidate selection, we train a
trajectory-conditioned critic to predict static and dynamic collision
risks from visual, pedestrian, goal, and candidate-trajectory
features. Static-risk targets are derived from obstacle clearance in
simulator maps, while dynamic-risk targets are constructed from
time-aligned pedestrian distances. Both risk heads are trained using
Smooth-$L_1$ regression.
For candidate trajectory $j$, let
$\hat c_{\mathrm{sta},j}$ and $\hat c_{\mathrm{dyn},j}$ denote the
predicted static and dynamic risks. We combine them with goal progress
$P_j$, smoothness cost $S_j$, and temporal inconsistency $C_j$:
\begin{equation}
V_j
=
-w_s\hat c_{\mathrm{sta},j}
-w_d\hat c_{\mathrm{dyn},j}
+w_gP_j
-w_mS_j
-w_cC_j.
\label{eq:critic_selection}
\end{equation}
The final trajectory is selected as
$\tau_{j^\ast}$, where $j^\ast=\arg\max_j V_j$.

\subsection{Ranger-GRPO}
\label{sec:flow_grpo}
Supervised flow matching tends to favor dominant demonstration
behaviors, limiting the exploration of alternative maneuvers
in dynamic human--robot interactions.
Building on Flow-GRPO~\cite{liu2025flowgrpo}, we propose
Ranger-GRPO, which combines skill-energy-guided SDE sampling
with group-relative policy optimization to discover
high-reward trajectories and distill them into an unguided
flow policy. The optimization process is illustrated in
Fig.~\ref{fig:ranger_framework} (right).

\noindent\textbf{Skill-Energy Guidance.}
For each skill $s$, privileged training information defines
a differentiable energy $\mathcal E_s(\boldsymbol\tau)$
that captures the corresponding social interaction objective.
Following gradient-based guidance for flow
matching~\cite{feng2025guidance},
we estimate the clean action by integrating the frozen
SFT velocity field from the current sampling state
$\mathbf x_t$ at time $t$ toward $t=0$:
\begin{equation}
\frac{\mathrm d\mathbf x_\xi^{\mathrm{ref}}}{\mathrm d\xi}
=
\mathbf v_{\mathrm{ref}}
(\mathbf x_\xi^{\mathrm{ref}},\xi,\mathbf o_k).
\end{equation}
Here, $\mathbf x_\xi^{\mathrm{ref}}$ denotes the reference
ODE state at integration time $\xi$, initialized from
$\mathbf x_t$.
Euler integration yields the clean-action estimate
$\hat{\mathbf a}_0$, which is mapped to a physical
trajectory $\hat{\boldsymbol\tau}=\mathcal T(\hat{\mathbf a}_0)$.
The guidance direction is then computed as:
\begin{equation}
\mathbf u_g(\mathbf x_t,t)
=
\nabla_{\mathbf x_t}
\mathcal E_s(\hat{\boldsymbol\tau}).
\label{eq:guidance_direction}
\end{equation}
Gradients propagate through the reference integration
to $\mathbf x_t$, while the SFT parameters remain frozen.

\noindent\textbf{Guided SDE Sampling.}
We introduce stochasticity through an SDE with drift
$\mathbf b_\theta
=\mathbf v_\theta+
\frac{\sigma^2(t)}{2t}
[\mathbf x_t+(1-t)\mathbf v_\theta]$.
Skill-energy guidance modifies the dynamics as:
\begin{equation}
\mathrm d\mathbf x_t
=
\left(
\mathbf b_\theta+
\gamma(t)\mathbf u_g(\mathbf x_t,t)
\right)\mathrm dt
+\sigma(t)\mathrm d\bar{\mathbf W}_t,
\label{eq:guided_flow_sde}
\end{equation}
where $\gamma(t)$ controls the guidance strength
and $\bar{\mathbf W}_t$ denotes reverse-time Brownian motion.
On the reverse-time grid $1=t_0>\cdots>t_N=0$,
where $\mathbf x_n$ denotes the state at time $t_n$
and $h_n=t_n-t_{n+1}>0$ is the step size,
Euler--Maruyama yields the unguided transition
$p^0_{\theta,n}
=\mathcal N(\boldsymbol\mu^0_{\theta,n},
\sigma_n^2\mathbf I_d)$,
where
$\boldsymbol\mu^0_{\theta,n}
=\mathbf x_n-h_n\mathbf b_{\theta,n}$
and $\sigma_n^2=\sigma^2(t_n)h_n$.
Guidance shifts the transition mean by
$\boldsymbol\Delta_n
=-h_n\gamma(t_n)\mathbf u_g(\mathbf x_n,t_n)$,
giving the guided sampling distribution:
\begin{equation}
q^g_{\mathrm{old},n}
=
\mathcal N\!\left(
\boldsymbol\mu^0_{\mathrm{old},n}
+\boldsymbol\Delta_n,
\sigma_n^2\mathbf I_d
\right).
\label{eq:guided_gaussian_policy}
\end{equation}
At each training round, we collect fresh observations
through closed-loop simulation and construct skill-balanced
batches. For each observation $\mathbf o_k$ assigned to
skill $s\in\mathcal S$, the old policy samples $G$
candidate trajectories $\{\boldsymbol\tau_{k,j}\}_{j=1}^{G}$
using the guided SDE.
Each guided transition is evaluated under the
current and old unguided policies to compute:
\begin{equation}
\rho_{k,j,n}
=
\frac{
p^0_{\theta,n}
(\mathbf x_{n+1}\mid\mathbf x_n,\mathbf o_k)
}{
p^0_{\mathrm{old},n}
(\mathbf x_{n+1}\mid\mathbf x_n,\mathbf o_k)
}.
\label{eq:unguided_likelihood_ratio}
\end{equation}

\noindent\textbf{Skill Learning by GRPO.}
Each candidate is evaluated using the corresponding skill
reward,
$R_{k,j}=R_s(\boldsymbol\tau_{k,j})$,
computed over all $H$ predicted steps.
The group-relative advantage is
$A_{k,j}=(R_{k,j}-\bar R_k)/
(\sigma_{R,k}+\varepsilon_R)$,
where $\bar R_k$ and $\sigma_{R,k}$ are the mean and standard
deviation of the $G$ rewards for the same observation.
This observation-wise normalization reduces sensitivity to reward-scale
differences across skills while preserving the ranking
of candidate trajectories within each group.
The trajectory-level advantage $A_{k,j}$ is assigned to all stochastic sampling transitions of candidate $j$.
Ranger-GRPO optimizes a shared policy across skills:
\begin{equation}
\mathcal J(\theta)=\mathbb E_{s,k,j,n}\!\Big[
\mathcal C(\rho_{k,j,n},A_{k,j})
-\beta_{\mathrm{KL}}D_{\mathrm{KL}}\!\left(
p^0_{\theta,n}\Vert p^0_{\mathrm{ref},n}
\right)\Big].
\label{eq:guided_grpo_objective}
\end{equation}
The expectation averages over skill-balanced observations, candidate trajectories, and stochastic transitions sampled from the guided old policy.
Following the clipped GRPO objective~\cite{shao2024deepseekmath},
the surrogate is
$\mathcal C(\rho,A)=
\min[\rho A,
\operatorname{clip}(\rho,1-\varepsilon_{\mathrm{clip}},
1+\varepsilon_{\mathrm{clip}})A]$.
The KL term constrains the refined policy relative to the
frozen SFT reference, with both transition distributions
conditioned on the same sampled state and observation.
All skills update the same policy parameters $\theta$.
The skill index selects the training reward and guidance
energy, but is not provided as a policy input.
Deployment uses only the unguided ODE policy without
privileged social information.
The complete training procedure is provided in the Appendix.

\begin{figure*}[t]
    \centering
    \captionsetup{skip=2pt}
    \includegraphics[width=\textwidth]{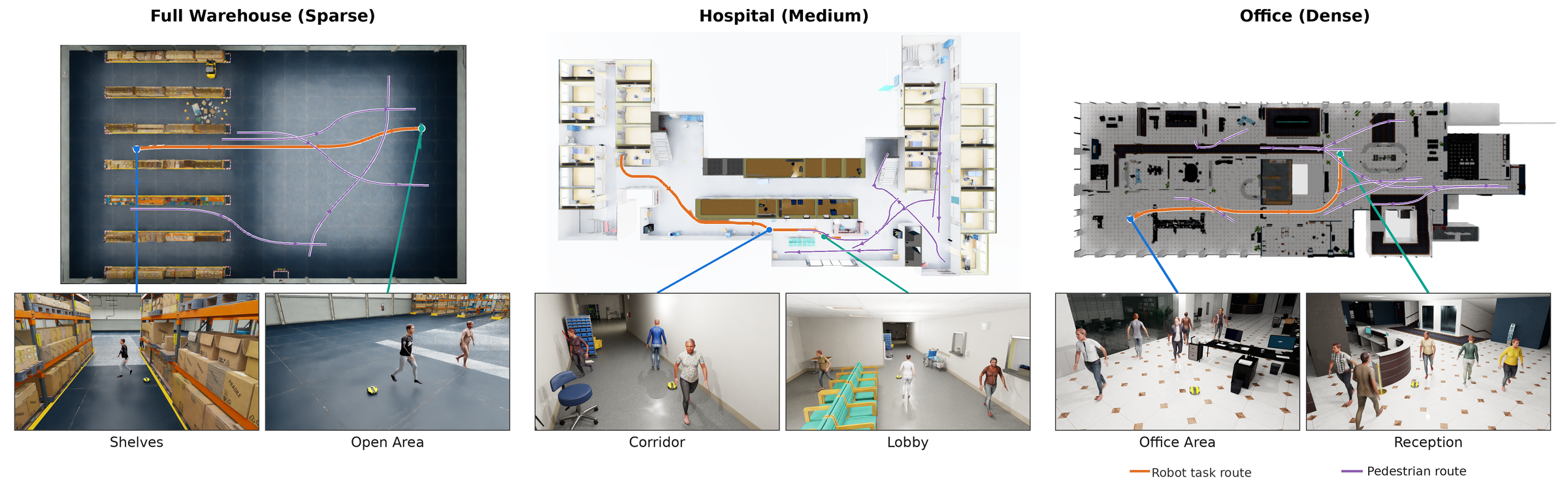}
    \caption{DynBench closed-loop evaluation environments with diverse layouts and sparse, medium, and dense pedestrian populations.}
    \label{fig:dynbench_scenes}
\end{figure*}

\begin{table*}[t]
\centering
\setlength{\tabcolsep}{3pt}
\resizebox{\textwidth}{!}{%
\begin{tabular}{l*{16}{c}}
\toprule
\multirow{2}{*}{Method}
& \multicolumn{4}{c}{Full Warehouse (Sparse)}
& \multicolumn{4}{c}{Hospital (Medium)}
& \multicolumn{4}{c}{Office (Dense)}
& \multicolumn{4}{c}{Average}\\
\cmidrule(lr){2-5}\cmidrule(lr){6-9}\cmidrule(lr){10-13}\cmidrule(lr){14-17}
& SR $\uparrow$ & SPL $\uparrow$ & MinD $\uparrow$ & CR $\downarrow$
& SR $\uparrow$ & SPL $\uparrow$ & MinD $\uparrow$ & CR $\downarrow$
& SR $\uparrow$ & SPL $\uparrow$ & MinD $\uparrow$ & CR $\downarrow$
& SR $\uparrow$ & SPL $\uparrow$ & MinD $\uparrow$ & CR $\downarrow$\\
\midrule
\rowcolor{black!7}
\multicolumn{17}{l}{\textcolor{black!55}{\textbf{Pre-trained Policies}}}\\
iPlanner~\cite{yang2023iplanner} & 65.5 & 64.27 & \textbf{\underline{2.162}} & \underline{7.5} & 17.0 & 16.74 & \textbf{\underline{1.408}} & \underline{27.5} & 19.5 & 19.30 & \textbf{\underline{1.617}} & \textbf{\underline{14.5}} & 34.00 & 33.44 & \textbf{\underline{1.729}} & \underline{16.50}\\
ViPlanner~\cite{roth2024viplanner} & 69.5 & \underline{68.02} & 2.145 & 8.0 & 33.0 & 32.03 & 1.255 & 33.5 & 31.0 & 30.49 & 1.469 & 21.0 & 44.50 & 43.51 & 1.623 & 20.83\\
NavDP~\cite{cai2026navdp}
& 65.5 & 58.65 & 2.104 & 12.5
& 42.5 & 39.92 & 0.981 & 39.0
& 41.0 & 38.31 & 1.040 & 34.0
& 49.67 & 45.63 & 1.375 & 28.50\\
SFT
& \underline{73.0} & 67.46 & 2.037 & 11.0
& \underline{53.0} & \underline{48.67} & 1.001 & 30.5
& \underline{46.5} & \underline{43.35} & 1.067 & 33.5
& \underline{57.50} & \underline{53.16} & 1.368 & 25.00\\
\midrule
\rowcolor{black!7}
\multicolumn{17}{l}{\textcolor{black!55}{\textbf{Post-trained Policies}}}\\
FLUX~\cite{gong2026flux}
& 64.0 & 58.39 & 2.033 & 15.5
& 38.0 & 36.04 & 1.072 & 38.5
& 44.0 & 41.49 & 1.148 & 31.0
& 48.67 & 45.31 & 1.418 & 28.33\\
X-NavDP~\cite{yang2026xnavdp}
& 74.0 & 62.75 & 2.014 & 9.5
& 41.0 & 37.10 & 0.987 & 38.5
& 44.0 & 39.25 & 1.049 & 36.0
& 53.00 & 46.37 & 1.350 & 28.00\\
NavOL~\cite{wei2026navol}
& \textbf{\underline{77.0}} & 66.78 & 1.600 & 17.0
& 40.0 & 37.04 & 0.982 & 42.5
& 39.5 & 36.76 & 1.053 & 30.5
& 52.17 & 46.86 & 1.212 & 30.00\\
\textbf{\method{}} & 76.0 & \textbf{\underline{68.35}} & \underline{2.127} & \textbf{\underline{7.0}} & \textbf{\underline{59.5}} & \textbf{\underline{52.17}} & \underline{1.174} & \textbf{\underline{17.0}} & \textbf{\underline{54.5}} & \textbf{\underline{48.40}} & \underline{1.187} & \underline{20.0} & \textbf{\underline{63.33}} & \textbf{\underline{56.31}} & \underline{1.496} & \textbf{\underline{14.67}}\\
\bottomrule
\end{tabular}%
}
\caption{Closed-loop navigation performance across the three DynBench environments under matched evaluation settings. \underline{Underlined} values indicate the best result within each pre-trained or post-trained group, while \textbf{bold} values indicate the best overall result.}
\label{tab:dynamic_navigation}
\vspace{-3mm}
\end{table*}
\section{Experiments}

\begin{figure*}[t]
    \centering
    \includegraphics[width=\textwidth]{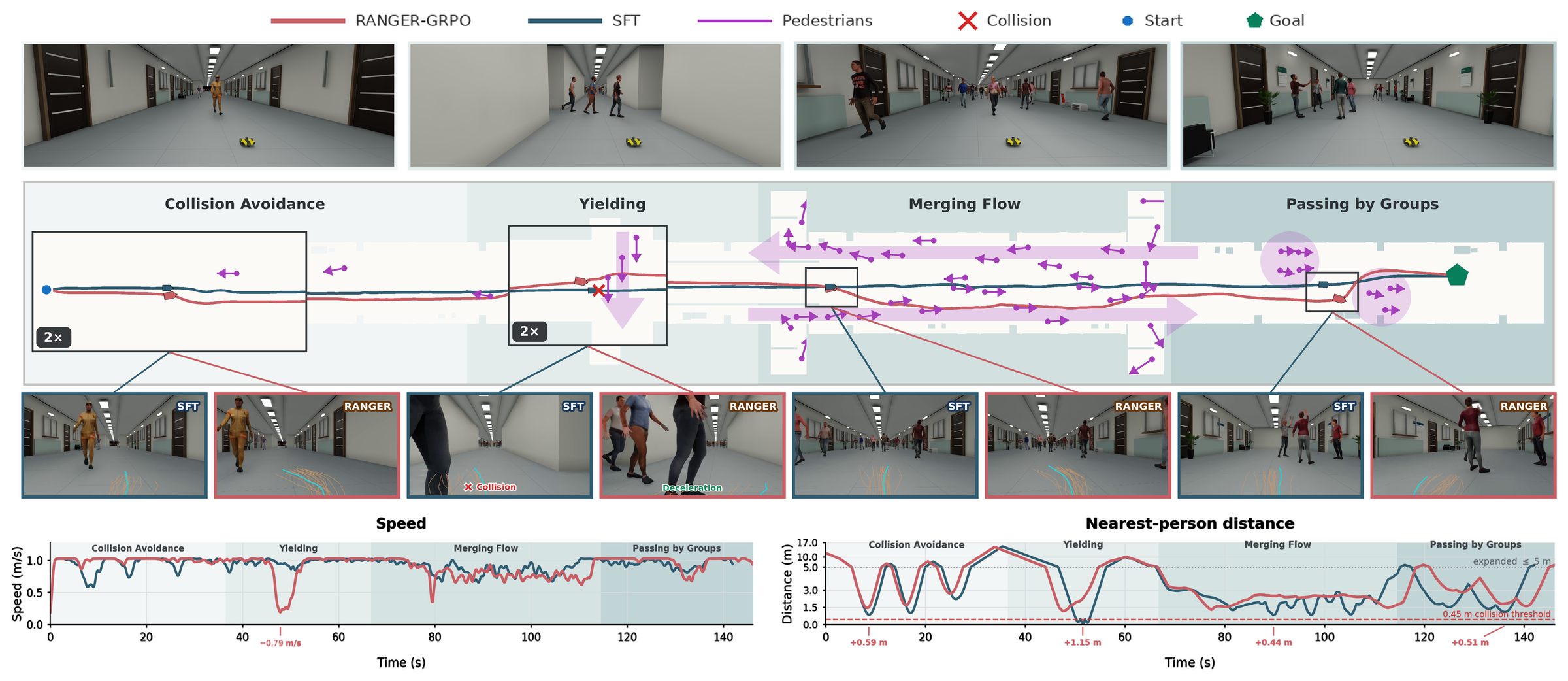}
    \caption{
    Closed-loop comparison of a navigation policy before (SFT)
    and after Ranger-GRPO across four consecutive scenarios:
    \emph{Collision Avoidance}, \emph{Yielding},
    \emph{Merging Flow}, and \emph{Passing by Groups}.
    The top row shows third-person views, while the middle rows
    compare map trajectories and first-person observations.
    The bottom row plots robot speed and nearest-person
    distance throughout the evaluation.
    }
    \label{fig:task_grpo_qualitative}
    \vspace{-3mm}
\end{figure*}

This section aims to address the following research questions:
\textbf{1)} Can \method{} improve closed-loop navigation performance in dynamic environments compared with existing visual navigation methods?
\textbf{2)} Can a single policy acquire diverse social navigation skills across different interaction scenarios?
\textbf{3)} Does Ranger-GRPO improve the exploration and learning of high-reward behaviors compared with existing reinforcement-learning paradigms for navigation policy post-training?
\textbf{4)} How do the individual components contribute to the overall navigation performance?
\textbf{5)} Can the learned policy generalize to real-world human-robot interactions?

\subsection{Setup and Protocols}

\noindent\textbf{SFT Details.}
The policy takes eight RGB frames sampled at $0.32$\,s intervals,
one depth frame, and a point goal, and predicts 24 trajectory steps
at $0.16$\,s intervals, corresponding to a $3.84$\,s prediction
horizon. The point goal and predicted trajectory are expressed in the
camera coordinate frame at the observation time.
We initialize the navigation policy from pretrained NavDP~\cite{cai2026navdp} weights.
The supervised training pipeline consists of two stages:
(1) train the pedestrian queries, prediction heads, and
RGB-D aggregation module while keeping the flow generator frozen;
and (2) jointly train the flow generator, trainable perception
modules, and trajectory critic.
The RGB-D ViT backbones remain frozen throughout training.


\noindent\textbf{Post-Training Details.}
We construct over 20 simulation assets with diverse visual
appearances for training four skills: \emph{Collision Avoidance},
\emph{Yielding}, \emph{Merging Flow}, and \emph{Passing by Groups}.
We initialize the policy from the SFT checkpoint.
During post-training, the current policy interacts with
the simulator to collect fresh observations online,
from which guided SDE sampling generates
candidate trajectories for reinforcement learning.
Across 400 rounds of online simulation, we collect
128 observations per round, equally distributed across
the four skills, and generate 16 trajectories per observation,
totaling 819,200 candidate trajectories.
Skill-specific guidance, rewards, and implementation
details are provided in the appendix.

\noindent\textbf{Benchmarks, Baselines and Metrics.}
We evaluate \method{} using a closed-loop interactive protocol built on
DynBench~\cite{gong2026flux} across Full Warehouse, Hospital, and Office,
with progressively increasing pedestrian density, as illustrated in
Fig.~\ref{fig:dynbench_scenes}.
These environments are distinct from our SFT and post-training data-generation assets.
For each evaluation episode, we augment the original benchmark
setting with dynamic robot--pedestrian interactions, where
pedestrians follow randomized routes and initial configurations and encounter the
robot along or across its path. This setup enables closed-loop
evaluation of navigation performance under active human-robot
interaction. An episode is considered unsuccessful if the robot
collides with a pedestrian or fails to reach the goal within the time limit.

Under the same evaluation protocol and matched controller settings,
we compare against pretrained navigation baselines, including
iPlanner~\cite{yang2023iplanner},
ViPlanner~\cite{roth2024viplanner},
NavDP~\cite{cai2026navdp}, and
our SFT policy used to initialize \method{}.
Post-trained baselines include
X-NavDP~\cite{yang2026xnavdp},
NavOL~\cite{wei2026navol}, and
FLUX~\cite{gong2026flux}.

For closed-loop evaluation, we report success rate (\textbf{SR}, \%),
success weighted by path length (\textbf{SPL}, \%), pedestrian
collision rate (\textbf{CR}, \%), and the mean of per-episode minimum
planar robot--pedestrian distances (\textbf{MinD}, m).
SPL averages $S d_0/\max(d_0,L)$ over episodes and is reported as
a percentage, where $S=1$ for success and $0$ otherwise, $d_0$ is
the initial Euclidean goal distance, and $L$ is the executed path
length. Table~\ref{tab:dynamic_navigation} reports results from 200 episodes
per scene, totaling 600 episodes per method.

\subsection{Main Results}


\noindent\textbf{Quantitative Results.}
Averaged over the three DynBench environments,
RANGER achieves a success rate of $63.33\%$ and an SPL of
$56.31\%$, outperforming the strongest external baselines
(X-NavDP for SR and NavOL for SPL)
by $10.33$ and $9.45$ percentage points, respectively, while
achieving the lowest average CR of $14.67\%$.
It also attains the largest average MinD among
post-trained methods ($1.496$\,m).
The gains are particularly pronounced in the
more crowded Hospital and Office environments, where RANGER
achieves the highest SR and SPL across all compared methods,
together with the lowest CR among post-trained methods.
These results support the effectiveness of Ranger-GRPO in
improving navigation success and pedestrian safety under
frequent human-robot interactions.

\noindent\textbf{Qualitative Comparison.}
We evaluate a single post-trained RANGER policy across four
consecutive interaction scenarios in a continuous closed-loop
rollout, using its SFT initialization as the baseline.
Fig.~\ref{fig:task_grpo_qualitative} compares their behaviors.
In \emph{Collision Avoidance}, RANGER starts lateral
avoidance earlier and passes the oncoming pedestrian with
a smoother recovery toward the goal.
In \emph{Yielding}, the SFT policy enters the crossing
region and collides, whereas RANGER slows down before
the interaction zone and accelerates after the pedestrian
has crossed.
In \emph{Merging Flow}, RANGER follows the surrounding
pedestrian flow, enabling smooth navigation through the crowd.
In \emph{Passing by Groups}, RANGER takes a wider detour
around the conversational group instead of cutting through
the occupied social space.
These consecutive interactions demonstrate how a single
RANGER policy adapts its spatial and temporal responses
to different social situations without skill labels or
policy switching.

\subsection{Ablation Studies}


\begin{table}[t]
\centering
\footnotesize
\setlength{\tabcolsep}{2pt}
\begin{tabular*}{\columnwidth}{@{\extracolsep{\fill}}lrrrr@{}}
\toprule
Variant & SR $\uparrow$ & SPL $\uparrow$
& MinD $\uparrow$ & CR $\downarrow$\\
\midrule
\rowcolor{black!7}
\multicolumn{5}{l}{\textcolor{black!55}{\textbf{Component Ablations}}}\\
SFT w/o pedestrian supervision
& 53.67 & 49.62 & 1.359 & 27.67\\
SFT w/o trajectory critic
& 54.33 & 50.48 & 1.363 & 25.67\\
SFT
& 57.50 & 53.16 & 1.368 & 25.00\\
\midrule
\rowcolor{black!7}
\multicolumn{5}{l}{\textcolor{black!55}{\textbf{Post-training Method Comparison}}}\\
SFT + Flow-GRPO & 60.00 & 47.26 & 1.453 & 15.67\\
SFT + RW & 58.00 & 52.01 & 1.473 & 20.00\\
\midrule
\textbf{\method{}}
& \textbf{63.33} & \textbf{56.31} & \textbf{1.496} & \textbf{14.67}\\
\bottomrule
\end{tabular*}
\caption{Closed-loop component ablations and post-training comparisons
on DynBench.}
\vspace{-3mm}
\label{tab:ablation}
\end{table}
\noindent\textbf{Component Ablations.}
Table~\ref{tab:ablation} evaluates the contributions of
pedestrian supervision, the trajectory critic, and
Ranger-GRPO post-training.
Removing pedestrian supervision from SFT reduces SR and SPL
by $3.83$ and $3.54$ percentage points, respectively, while
increasing CR by $2.67$ points.
Removing the trajectory critic reduces SR and SPL by $3.17$
and $2.68$ points,
supporting its contribution to navigation performance.
Applying Ranger-GRPO to the full SFT model further increases
SR and SPL by $5.83$ and $3.15$ percentage points,
respectively, reduces CR from $25.00\%$ to $14.67\%$,
and increases MinD from $1.368$\,m to $1.496$\,m.
These results support the contributions of both SFT components
and the additional gains from guided reinforcement post-training.

\begin{figure*}[t]
    \centering
    \includegraphics[width=\textwidth]{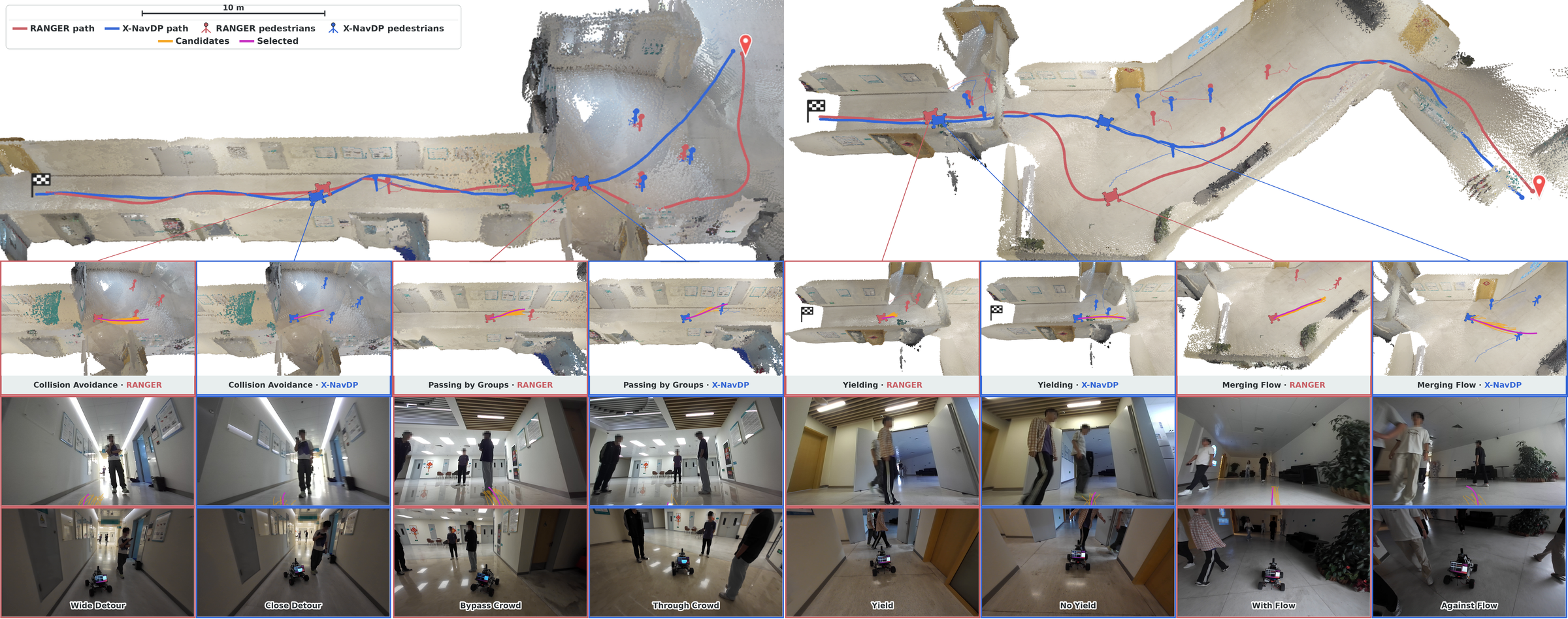}
    \caption{Real-world comparison between RANGER (red) and X-NavDP (blue)
    across four interaction scenarios. The top row shows the executed robot
    paths and pedestrian positions; the second row shows local maps with
    candidate and selected trajectories; and the bottom two rows show the
    corresponding first- and third-person views.}
    \label{fig:real_world_deployment}
\end{figure*}

\begin{table*}[t]
\centering
\setlength{\tabcolsep}{1.5pt}
\resizebox{\textwidth}{!}{%
\begin{tabular}{l*{20}{c}}
\toprule
\multirow{2}{*}{Method}
& \multicolumn{5}{c}{\emph{Collision Avoidance}}
& \multicolumn{5}{c}{\emph{Yielding}}
& \multicolumn{5}{c}{\emph{Merging Flow}}
& \multicolumn{5}{c}{\emph{Passing by Groups}}\\
\cmidrule(lr){2-6}\cmidrule(lr){7-11}\cmidrule(lr){12-16}\cmidrule(lr){17-21}
& Mean $\uparrow$ & Top-1 $\uparrow$ & P-Col $\downarrow$ & Prog. $\uparrow$ & Spread $\uparrow$
& Mean $\uparrow$ & Top-1 $\uparrow$ & P-Col $\downarrow$ & Prog. $\uparrow$ & Spread $\uparrow$
& Mean $\uparrow$ & Top-1 $\uparrow$ & P-Col $\downarrow$ & Prog. $\uparrow$ & Spread $\uparrow$
& Mean $\uparrow$ & Top-1 $\uparrow$ & P-Col $\downarrow$ & Prog. $\uparrow$ & Spread $\uparrow$\\
\midrule
SFT
& -1.121 & -0.749 & 13.26 & 2.883 & \textbf{0.163}
& -8.599 & -7.357 & 17.66 & \textbf{3.355} & \underline{0.098}
& 1.178 & 2.858 & 39.80 & 2.144 & \textbf{0.163}
& -2.567 & -2.220 & 27.46 & \textbf{2.425} & \textbf{0.157}\\
Flow-GRPO
& \underline{-0.955} & \underline{-0.741} & \underline{11.85} & \underline{3.056} & 0.126
& \textbf{-5.245} & \underline{-4.654} & \textbf{12.76} & \underline{2.588} & \underline{0.098}
& \underline{3.567} & \underline{4.716} & \underline{22.66} & \underline{2.301} & 0.133
& \underline{-2.059} & \underline{-1.785} & \textbf{13.24} & 2.235 & 0.137\\
RW
& -1.159 & -0.851 & \textbf{11.41} & 2.795 & 0.121
& -5.445 & -4.974 & 20.01 & 2.426 & 0.066
& 1.042 & 1.727 & 37.93 & 1.926 & 0.063
& -2.714 & -2.489 & 21.77 & 2.022 & 0.073\\
RANGER
& \textbf{-0.924} & \textbf{-0.712} & \underline{11.85} & \textbf{3.061} & \underline{0.134}
& \underline{-5.390} & \textbf{-4.600} & \underline{13.17} & 2.489 & \textbf{0.121}
& \textbf{3.684} & \textbf{4.981} & \textbf{21.47} & \textbf{2.351} & \underline{0.159}
& \textbf{-2.001} & \textbf{-1.719} & \underline{13.75} & \underline{2.311} & \underline{0.154}\\
\bottomrule
\end{tabular}}
\caption{Open-loop performance across four social navigation skills.
\textbf{Bold} and \underline{underline} indicate the best and
second-best results in each column, respectively.}
\label{tab:skill_openloop}
\vspace{-3mm}
\end{table*}

\noindent\textbf{Post-Training Method Comparison.}
Ranger-GRPO is compared with two representative post-training paradigms: Reward Weighting (RW)~\cite{zhang2026sidp,huang2026grace} for reward-weighted generative refinement and Flow-GRPO~\cite{liu2025flowgrpo,sheng2026beyond,chen2026socialnav} for likelihood-ratio-based policy optimization. All methods start from the same pedestrian-aware SFT policy and share identical rewards, sampling budgets, and training budgets. Open-loop evaluation uses 6,554 fixed validation observations: 500 for \emph{Collision Avoidance}, 1,105 for \emph{Yielding}, 2,249 for \emph{Merging Flow}, and 2,700 for \emph{Passing by Groups}, with no episode overlap with post-training data. Each method generates 16 trajectories per observation via unguided ODE sampling, with results averaged over five evaluations using distinct initial noise sets shared across methods. Evaluation metrics include the mean trajectory reward (\textbf{Mean}), the highest reward among 16 candidates averaged over observations (\textbf{Top-1}), the proportion of candidates colliding with recorded future pedestrian trajectories (\textbf{P-Col}, \%), goal progress measured by the reduction in goal distance (\textbf{Prog.}, m), and trajectory diversity measured by the mean pairwise waypoint distance across candidates, averaged over prediction steps and observations (\textbf{Spread}, m).

Table~\ref{tab:skill_openloop} shows that RANGER improves trajectory quality while retaining greater diversity than the other post-training methods. RANGER achieves the highest Top-1 reward across all four skills and the highest mean reward in three. Among the post-training methods, RANGER achieves the greatest progress in Collision Avoidance, Merging Flow, and Passing by Groups, as well as the largest trajectory spread across all four skills. RANGER achieves the lowest P-Col in Merging Flow, with collision rates close to those of Flow-GRPO in the other three skills.

The closed-loop results in Table~\ref{tab:ablation} further support the
effectiveness of guided post-training. Among the compared post-training
methods, RANGER achieves
the highest SR ($63.33\%$) and SPL ($56.31\%$), the largest MinD
($1.496$\,m), and the lowest CR ($14.67\%$).
Together, the open- and closed-loop results support skill-energy guidance
as a key mechanism for directing exploration toward high-reward
interaction trajectories and transferring the resulting behaviors
into a single unguided navigation policy.


\subsection{Real-World Experiments}

We compare two visual navigation policies in real-world
deployment: RANGER and the X-NavDP baseline.
Both are deployed on the same Scout Mini mobile robot equipped
with a ZED~2 RGB--D camera and a Livox MID-360 LiDAR.
FAST-LIO2~\cite{xu2022fastlio2} provides the robot state
estimate used to construct the point-goal input, while
both policies receive only RGB--D observations and the
point goal, without LiDAR measurements.
At each planning step, both methods sample 16 candidate
trajectories, select one using their respective
trajectory-selection mechanisms, and track it with a
model-predictive controller.
Throughout deployment, a single RANGER checkpoint performs
all four social interactions using only a final point goal
and onboard observations, without privileged pedestrian
states, skill labels, model switching, or online adaptation.

Fig.~\ref{fig:real_world_deployment} compares the two
policies across four representative real-world interactions.
RANGER passes around a pedestrian group, takes a wider
detour around an oncoming pedestrian, yields at a doorway,
and merges with the surrounding flow.
In the matched interactions, X-NavDP instead passes through
the group, takes a closer detour, proceeds without yielding,
and moves against the flow.
These results illustrate the transfer of diverse social
navigation skills learned in simulation to real-world
human environments.

\section{Conclusion}

We introduced RANGER, combining interaction-rich demonstrations, pedestrian-aware flow modeling, and guided reinforcement post-training for social navigation. Ranger-GRPO uses skill-energy-guided SDE sampling to refine an unguided policy through group-relative optimization. Experiments demonstrate improved navigation success and reduced pedestrian collisions over existing methods, while a single policy exhibits four social navigation skills. Real-robot deployment demonstrates these behaviors without fine-tuning or inference-time guidance. Future work will explore interaction-aware pedestrian prediction to model human responses to robot trajectories.


\clearpage
\endgroup
{\small\section*{LLM Usage Statement}

OpenAI ChatGPT was used as an assistive tool for language editing, restructuring, and polishing parts of the manuscript; reviewer-style feedback; literature search and organization; code assistance for implementation and data visualization; and limited discussion of experimental design and result interpretation. All AI-assisted content was reviewed and verified by the authors. All experimental data, evaluations, and final scientific claims and conclusions were produced, checked, and approved by the authors, who take full responsibility for the submitted content.

}

\clearpage
\appendix
\maketitlesupplementary
\phantomsection\label{supplementary:start}

\numberwithin{equation}{section}

\pdfbookmark[1]{Supplementary Contents}{supplementary_contents}
\noindent{\large\bfseries Contents}\par\medskip
\begingroup
\small
\newcommand{\suppcontentsline}[1]{%
    \smallskip\noindent
    \hyperref[#1]{\ref*{#1}\quad\nameref*{#1}}%
    \nobreak\dotfill\hyperref[#1]{\pageref*{#1}}\par}
\newcommand{\suppcontentssubline}[3][1.4em]{%
    \noindent\hspace*{#1}\hyperref[#2]{#3}%
    \nobreak\dotfill\hyperref[#2]{\pageref*{#2}}\par}
\suppcontentsline{app:data_generation}
\suppcontentssubline{app:contents_maskedmimic}{MaskedMimic Training and Deployment}
\suppcontentssubline{app:contents_experts}{Expert Design and Demonstration Collection}
\suppcontentsline{app:sft_training_curves}
\suppcontentssubline{app:contents_training_stages}{Training Stages}
\suppcontentssubline{app:contents_critic}{Trajectory Critic}
\suppcontentsline{app:ranger_guided_distillation}
\suppcontentssubline{app:ranger_fm_score}{Flow Matching}
\suppcontentssubline{app:ranger_flow_sde}{Stochastic Flow}
\suppcontentssubline{app:ranger_guided_gaussians}{Guided Sampling}
\suppcontentssubline{app:ranger_unguided_likelihood}{Unguided Likelihood}
\suppcontentssubline{app:ranger_kl_projection}{KL Projection}
\suppcontentssubline{app:ranger_velocity_refinement}{Flow Refinement}
\suppcontentssubline{app:ranger_reward_selective}{GRPO Refinement}
\suppcontentsline{sec:appendix_algorithm}
\suppcontentssubline{app:contents_training_hyperparameters}{Training Hyperparameters}
\suppcontentssubline{app:contents_noise_schedule}{Noise Schedule}
\suppcontentssubline{app:contents_guidance_implementation}{Guidance Implementation}
\suppcontentsline{app:skill_rewards}
\suppcontentssubline{app:common_reward}{Shared Motion and Safety Reward}
\suppcontentssubline{app:headon_reward}{Collision Avoidance}
\suppcontentssubline{app:yield_reward}{Yielding}
\suppcontentssubline{app:follow_reward}{Merging Flow}
\suppcontentssubline{app:conversation_reward}{Passing by Groups}
\suppcontentsline{app:guidance_qualitative}
\suppcontentssubline{app:contents_guided_sde}{Guided SDE Exploration}
\suppcontentssubline{app:contents_trajectory_generation}{Guidance Effects on Trajectory Generation}
\suppcontentssubline{app:contents_guidance_ablation}{Guidance-Strength Ablation}
\suppcontentsline{app:qualitative_diagnostics}
\suppcontentssubline{app:contents_quantitative}{Quantitative Analysis}
\suppcontentssubline{app:contents_qualitative}{Qualitative Analysis of Learned Behaviors}
\suppcontentsline{app:classroom_deployment}
\suppcontentssubline{app:contents_deployment_setup}{Deployment Setup}
\suppcontentssubline{app:contents_behavioral_analysis}{Behavioral Analysis}
\suppcontentsline{app:dynbench_qualitative}
\suppcontentssubline{app:contents_scene_design}{Scene and Pedestrian Design}
\suppcontentssubline{app:contents_robot_protocol}{Robot Deployment and Evaluation Protocol}
\suppcontentssubline{app:contents_dynbench_comparisons}{Qualitative Comparisons}
\suppcontentsline{app:limitations}
\endgroup
\medskip

\begin{figure*}[t]
    \centering
    \includegraphics[width=\textwidth]{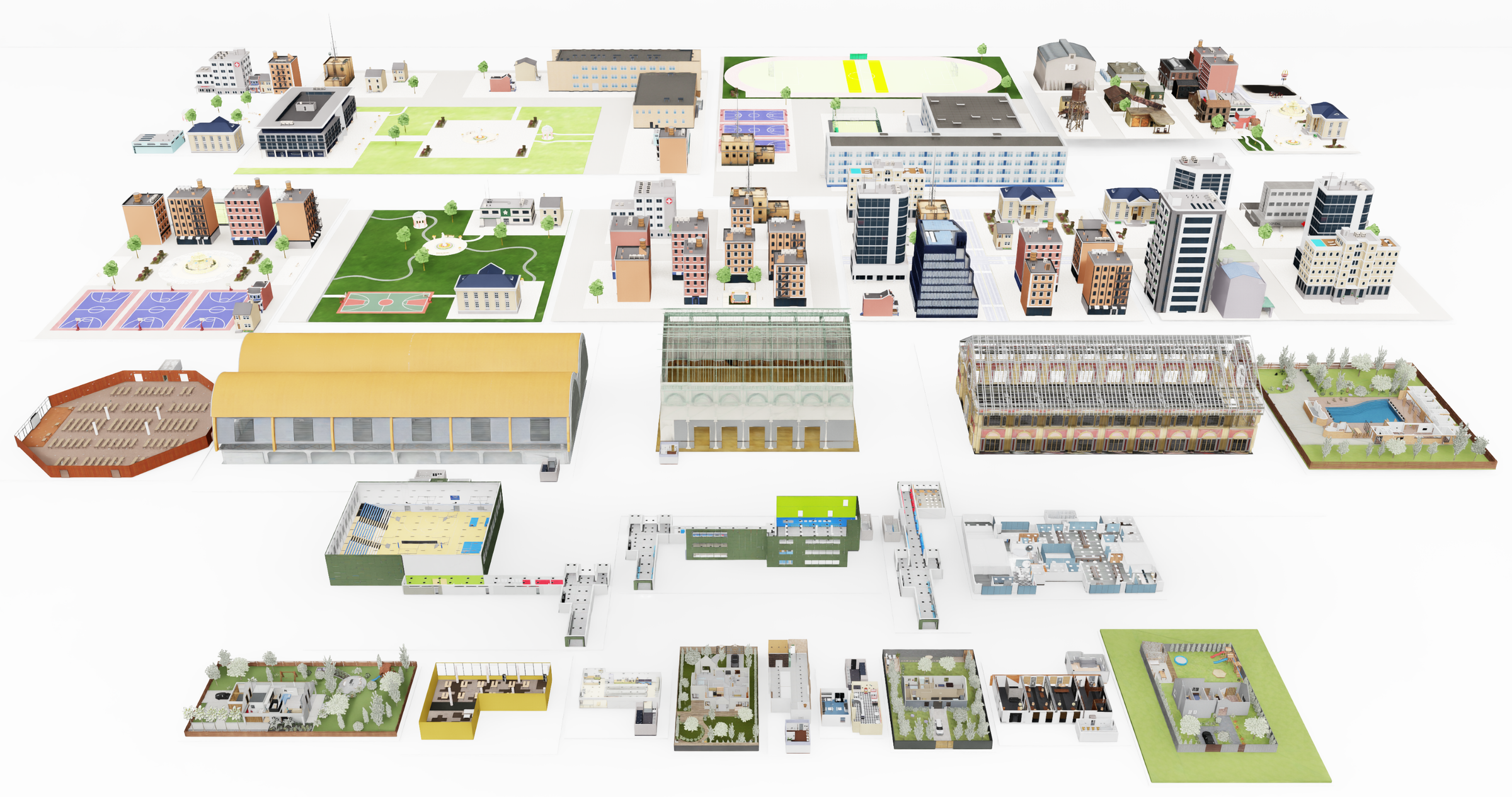}
    \caption{Overview of the diverse indoor and outdoor simulation
    environments used by our interaction-rich demonstration collection
    pipeline. The environment suite spans residential, educational,
    healthcare, commercial, recreational, and urban layouts with varied
    geometry, scale, and visual appearance.}
    \label{fig:isaacnav_environment_overview}
\end{figure*}

\section{Data Generation Details}
\label{app:data_generation}

\noindent\phantomsection\label{app:contents_maskedmimic}%
\textbf{MaskedMimic Training and Deployment.}
We employ the SMPL-based 
MaskedMimic~\cite{tessler2024maskedmimic}
controller to generate physically plausible
whole-body pedestrian motion from sparse constraints.
The MaskedMimic training procedure
consists of two stages.

First, a fully constrained (FC) teacher policy
$\pi_{\mathrm{FC}}^{\mathrm{char}}$ is trained
through reinforcement learning to track
reference body motions.
Its objective is:
\begin{equation}
\mathcal J_{\mathrm{FC}}^{\mathrm{char}}
=
\mathbb E_{\pi_{\mathrm{FC}}^{\mathrm{char}}}
\left[
\sum_{\ell=0}^{T_{\mathrm{char}}-1}
\gamma_{\mathrm{char}}^\ell
r_{i,\ell}^{\mathrm{track}}
\right],
\label{eq:maskedmimic_teacher}
\end{equation}
where $r_{i,\ell}^{\mathrm{track}}$
combines joint-position and rotation tracking,
root-height and velocity matching,
and an energy penalty.

Second, the teacher is distilled into a partially
constrained student $\pi_{\mathrm{char}}$
through DAgger-style behavioral cloning.
Random masking transforms full motion targets
$\mathbf g_{i,k}^{\mathrm{full}}$
into sparse conditions
$\mathbf g_{i,k}^{\mathrm{sparse}}$.
The student employs a conditional VAE
with a learned prior $p_{i,k}^{\mathrm{char}}$,
a privileged posterior $q_{i,k}^{\mathrm{char}}$,
and a decoder $\mathcal D^{\mathrm{char}}$.
The prior conditions on the body state and
sparse targets, while the posterior additionally
accesses full motion targets and predicts
a residual correction to the prior mean.
Following~\cite{tessler2024maskedmimic},
the core distillation objective is:
\begin{equation}
\begin{aligned}
\mathcal L_{\mathrm{char}}
=\mathbb E\Big[
&-\log\mathcal D^{\mathrm{char}}
(\mathbf u_{i,k}^{\mathrm{FC}}
\mid\mathbf s_{i,k}^{\mathrm{body}},
\boldsymbol\zeta_{i,k})\\
&+\lambda_{\mathrm{char}}
D_{\mathrm{KL}}
(q_{i,k}^{\mathrm{char}}
\|p_{i,k}^{\mathrm{char}})
\Big],
\end{aligned}
\label{eq:maskedmimic_student}
\end{equation}
where $\mathbf u_{i,k}^{\mathrm{FC}}$
is the teacher action and
$\boldsymbol\zeta_{i,k}$ is sampled from
the privileged posterior.
The expectation covers student-visited states,
teacher actions, and posterior latents.

During deployment, only the learned prior
and decoder are retained.
The student $\pi_{\mathrm{char}}$ in Eq.~\eqref{eq:pedestrian_generation}
samples a 64-dimensional latent vector as:
\begin{equation}
\boldsymbol\zeta_{i,k}
=
\boldsymbol\mu_{i,k}^{\mathrm{char}}
+
\boldsymbol\sigma_{i,k}^{\mathrm{char}}
\odot\mathbf e_i,
\quad
\mathbf e_i\sim\mathcal N(\mathbf 0,\mathbf I_{64}),
\label{eq:maskedmimic_latent}
\end{equation}
where the prior predicts the mean and standard
deviation. The noise vector $\mathbf e_i$
is sampled independently for each pedestrian
and held fixed throughout an episode,
while the latent distribution remains
state-dependent.
Sparse future pelvis positions and orientations
from the high-level navigation module,
together with the current body state and
motion history, condition the student to
generate 69-dimensional joint position targets.
These targets are executed by the Isaac Sim
PD controller, which converts them into torques.
The resulting body states are fed back into
$\pi_{\mathrm{char}}$ for closed-loop
whole-body motion generation.

\noindent\phantomsection\label{app:contents_experts}%
\textbf{Expert Design and Demonstration Collection.}
We generate diverse human--robot interaction demonstrations
using three complementary sources: a privileged PPO policy,
an Optimal Reciprocal Collision Avoidance (ORCA) planner,
and human teleoperation.

\begin{figure}[t]
    \centering
    \includegraphics[width=\columnwidth]{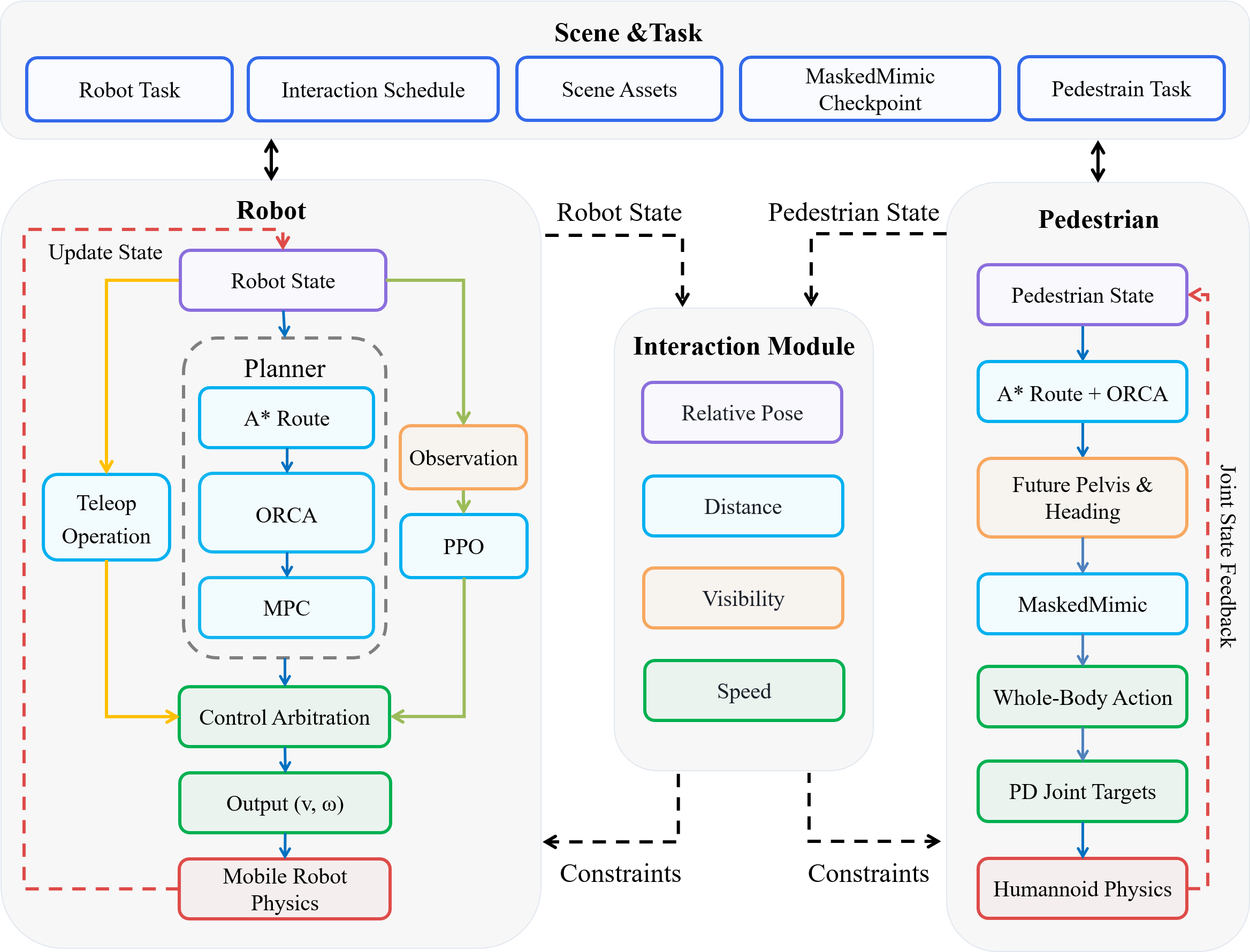}
    \caption{Closed-loop demonstration collection framework.
    The interaction schedule coordinates robot and pedestrian
    navigation tasks, while the interaction module provides
    state-dependent constraints to both agents during simulation.}
    \label{fig:demonstration_pipeline}
\end{figure}

\begin{figure*}[t]
    \centering
    \includegraphics[width=\textwidth]{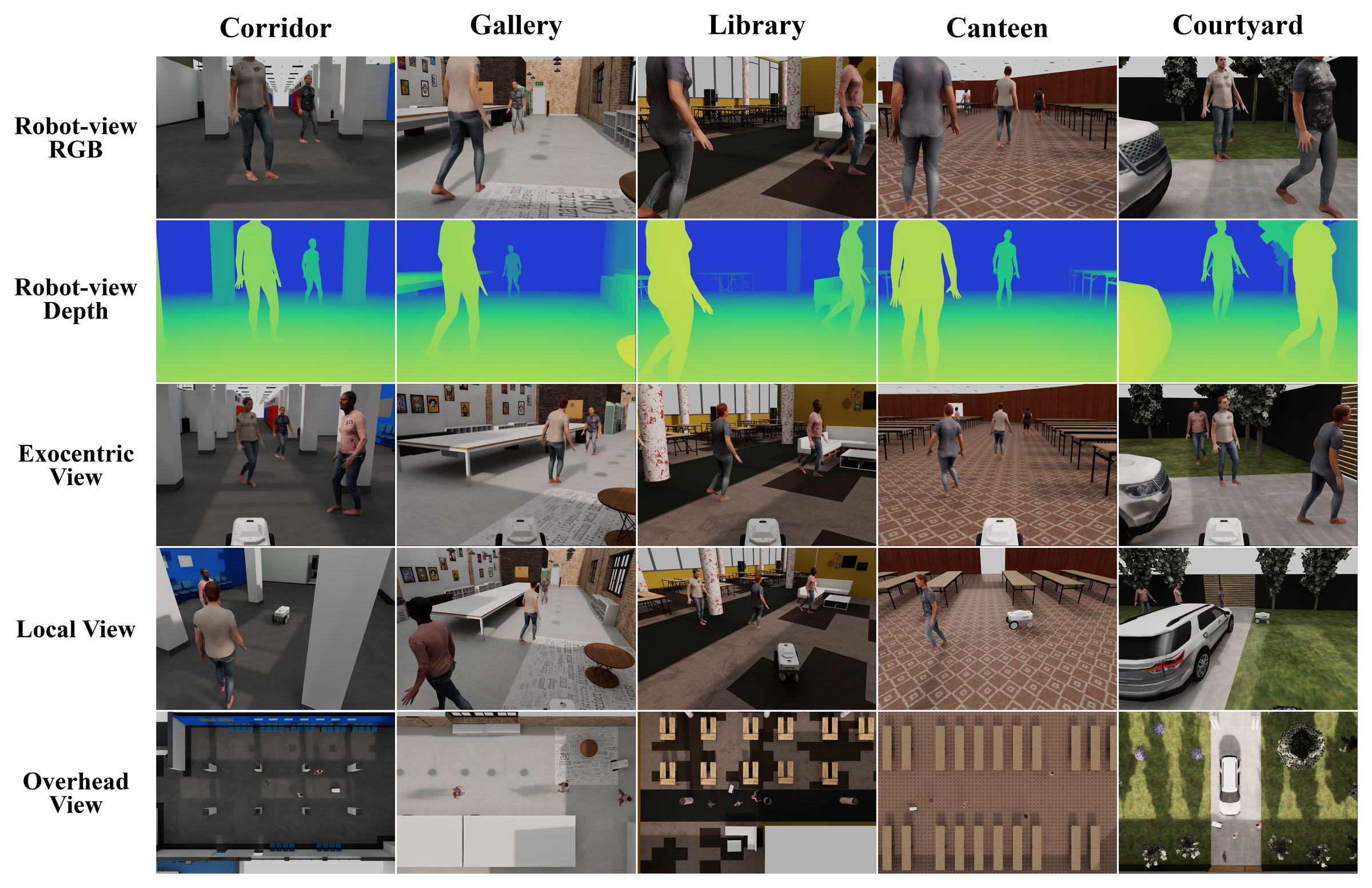}
    \caption{Representative scenes and observations from our interaction-rich
    data generation pipeline. Columns show five visually and geometrically
    diverse environments; rows show robot-view RGB and depth observations,
    together with exocentric, local, and overhead views of the corresponding
    robot--pedestrian interactions.}
    \label{fig:data_generation_examples}
\end{figure*}

The privileged policy is trained for goal-conditioned navigation
using PPO~\cite{schulman2017ppo} in NavIsaacLab.
It receives privileged simulator observations
$\mathbf{o}_t^{\mathrm{PPO}}=
(\mathbf{s}_t,\mathbf{g}_t,\mathbf{N}_t,\mathbf{M}_t,\mathbf{D}_t)$,
where $\mathbf{s}_t$ contains the robot's linear and angular
velocities, $\mathbf{g}_t$ represents the relative goal,
and $\mathbf{N}_t\in\mathbb{R}^{4\times5}$ encodes the
distances, relative bearings, and velocities of the four
nearest agents within a 4\,m radius.
The local occupancy map
$\mathbf{M}_t\in\mathbb{R}^{24\times24}$ covers an
$8\,\mathrm{m}\times8\,\mathrm{m}$ region, while
$\mathbf{D}_t$ provides depth observations.
Separate encoders process vector, occupancy-map, and depth
features, which are fused and passed to the actor--critic
networks.
The policy outputs linear and angular velocity commands
$\mathbf{a}_t=(v_t,\omega_t)$, with
$v_t\in[0,1]$\,m/s and
$\omega_t\in[-1,1]$\,rad/s.
The reward encourages goal progress, goal-directed motion,
collision avoidance, and smooth control, with additional
terminal rewards and penalties for goal reaching, collisions,
timeouts, and stagnation.
Training uses a discount factor of $0.99$,
a GAE parameter of $0.95$, a clipping ratio of $0.2$,
and a learning rate of $3\times10^{-4}$.
A goal curriculum progressively increases navigation
distance and initial heading deviation to expose
the policy to more challenging interactions.

To complement the learned policy, we use
ORCA~\cite{vandenberg2011orca} for reactive collision avoidance.
At each planning step, ORCA constructs velocity obstacles
from the relative positions and velocities of nearby agents
over a finite prediction horizon.
Reciprocal collision-avoidance constraints define a set
of admissible velocities, from which the planner selects
the velocity closest to the preferred goal-directed motion.
The resulting commands are continuously updated as
neighboring agents move, enabling collision avoidance
and local trajectory adjustments without a learned policy.
Human teleoperation further complements these automated
experts by capturing context-dependent navigation decisions
and interaction timing that may be underrepresented in
reactive or reward-driven policies.
Operators directly control the simulated robot through
linear and angular velocity commands, collecting
demonstrations of yielding to crossing pedestrians,
waiting in constrained passages, passing around
pedestrian groups, and recovering from blocked routes.

\begin{figure*}[t]
    \centering
    \includegraphics[width=\textwidth]{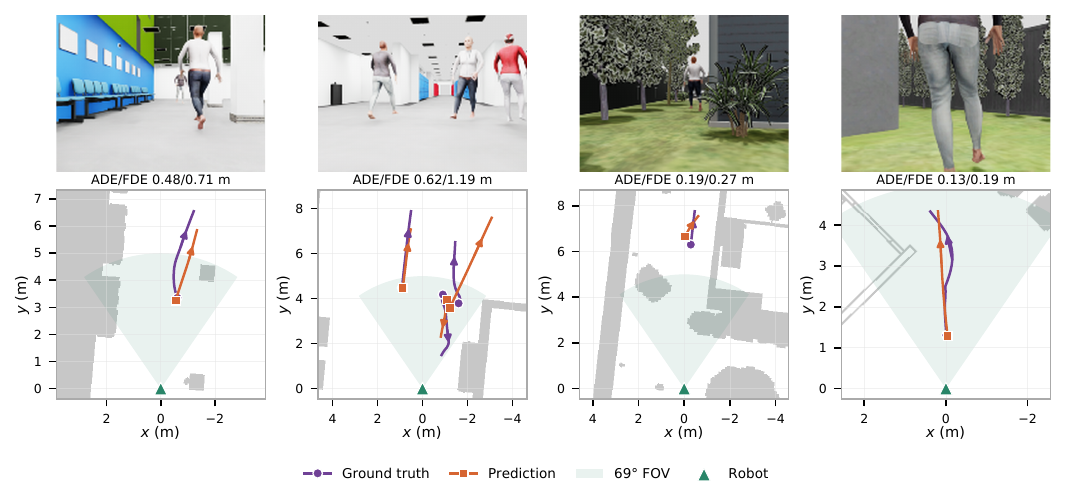}
    \caption{Qualitative pedestrian predictions on validation samples.
    Gray regions indicate occupied map cells, and shaded
    sectors illustrate the camera's horizontal field of view (FOV).}
    \label{fig:pedestrian_validation_examples}
\end{figure*}

\begin{figure}[t]
    \centering
    \includegraphics[width=\columnwidth]{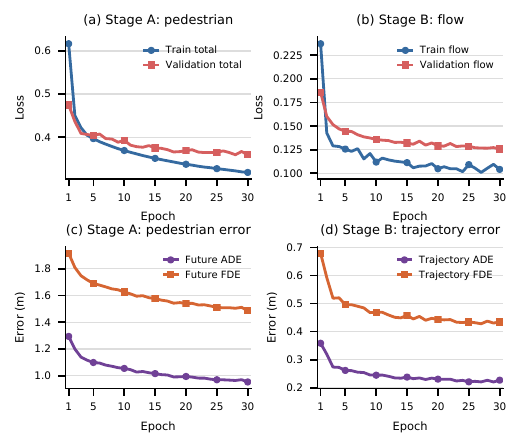}
    \caption{SFT optimization over 30 epochs per stage. The top row
    shows training and validation losses for pedestrian prediction
    in Stage A and flow matching in Stage B. The bottom row reports
    validation pedestrian future ADE/FDE and trajectory ADE/FDE.}
    \label{fig:sft_training_loss}
\end{figure}

Demonstrations from all three sources are collected through
a shared pipeline implemented in Isaac Sim, using simulated
Nova Carter mobile robots equipped with RGB--D cameras
across 50 indoor and outdoor
OmniGibson~\cite{li2022behavior} scenes, as summarized in
Fig.~\ref{fig:isaacnav_environment_overview}.
Figure~\ref{fig:demonstration_pipeline} illustrates the
closed-loop collection framework.
To encourage frequent and diverse human--robot interactions,
an interaction module coordinates robot and pedestrian
navigation tasks through spatial and temporal constraints.
Pedestrian routes are selected to intersect or overlap
with the robot route, while their initial positions,
motion directions, and arrival times are coordinated
to generate diverse encounters, including head-on passing,
crossing, and merging into pedestrian flows.
During simulation, the interaction module continuously
monitors the relative poses, distances, visibility, and
velocities of the robot and pedestrians, providing
interaction-dependent constraints to their respective
navigation modules.
Pedestrian motion targets are updated through ORCA and
converted into whole-body actions by MaskedMimic,
allowing robot and pedestrian behaviors to mutually
influence each other throughout the rollout.
We combine demonstration episodes from ORCA, PPO,
and human teleoperation at a ratio of $6:3:1$,
respectively.
All experts operate in closed loop, with synchronized
recording of RGB--D observations, robot poses, executed
actions, point goals, and pedestrian trajectories.
Simulator-derived geometry and pedestrian states are
retained as privileged training information but are not
provided to the final visual policy.
In total, we collect 94.05\,km of navigation trajectories.
As illustrated in Fig.~\ref{fig:data_generation_examples},
the resulting dataset captures diverse visual and geometric
conditions, complementary navigation strategies, and
temporally coherent human--robot interactions.

\section{SFT Training Details}
\label{app:sft_training_curves}

Our dataset contains 681,129 trajectory windows.
Within each OmniGibson scene, we split episodes into
90\% for training and 10\% for validation.
Each window contains eight historical
RGB frames sampled at $0.32\,\mathrm{s}$ intervals
(a $2.24\,\mathrm{s}$ history), the current depth image and
point goal, and 24 future action targets sampled at
$0.16\,\mathrm{s}$ intervals (a $3.84\,\mathrm{s}$ horizon).
It also includes current and 24-step future supervision
for up to four visible pedestrians.

\noindent\phantomsection\label{app:contents_training_stages}%
\textbf{Training Stages.}
The model is initialized from the original NavDP checkpoint.
The RGB and depth ViT backbones remain frozen throughout
training. Within the RGB--D fusion encoder, only the final
Transformer layer and output projection are updated,
while all preceding layers remain frozen.
Both training stages run for 30 epochs, with training
and validation loss curves shown in
Fig.~\ref{fig:sft_training_loss}.

In Stage A, we freeze the flow generator and train the
four learnable pedestrian queries, their cross-attention
and normalization layers, the position, motion, and
existence prediction heads, and the trainable RGB--D
fusion components.
For the pedestrian auxiliary loss in
Eq.~\eqref{eq:pedestrian_loss}, we set
$\lambda_{\mathrm{cur}}=\lambda_{\mathrm{delta}}
=\lambda_{\mathrm{exist}}=1$.
The current-position and future-displacement regression
terms use Smooth-$L_1$ losses, while the existence
classification term uses binary cross-entropy.
We use AdamW with a batch size of 368 and a weight
decay of $10^{-4}$.
The learning rates are $5\times10^{-5}$ for the
pedestrian modules and $5\times10^{-6}$ for the
trainable RGB--D fusion components.
We select the checkpoint with the lowest validation
pedestrian future ADE.
Fig.~\ref{fig:pedestrian_validation_examples}
shows qualitative pedestrian position and motion
predictions on the validation set.

In Stage B, we initialize from the selected Stage A
checkpoint and jointly optimize the flow generator,
pedestrian modules, and trajectory critic.
The flow and pedestrian modules are trained with
$\mathcal L_{\mathrm{FM}}+\mathcal L_{\mathrm{ped}}$.
The learning rates are $2\times10^{-5}$ for the
flow generator and trajectory critic,
$2\times10^{-6}$ for the pedestrian modules,
and $5\times10^{-7}$ for the trainable
RGB--D fusion components.

\noindent\phantomsection\label{app:contents_critic}%
\textbf{Trajectory Critic.}
The critic contains separate static- and dynamic-risk heads.
Static targets are computed from candidate clearance to simulator
obstacles, whereas dynamic targets use time-aligned distances to
future pedestrian trajectories. Both heads are trained with equally
weighted Smooth-$L_1$ regression losses. At deployment, the predicted
risks are combined with directly computed goal progress, trajectory
smoothness, and temporal continuity using
Eq.~\eqref{eq:critic_selection}. We set:
\begin{equation}
(w_s,w_d,w_g,w_m,w_c)=(1,3,1,0.1,1).
\label{eq:critic_selection_weights}
\end{equation}
Here, $P_j$ is the reduction in point-goal distance,
$S_j$ measures changes in successive planar and heading increments,
and $C_j$ is the mean waypoint deviation from the unexecuted portion
of the previously selected trajectory after temporal alignment.
We set $C_j=0$ at the first planning step. The larger dynamic-risk
weight prioritizes pedestrian avoidance, while the continuity and
smoothness terms suppress abrupt plan switching and oscillatory
motion.

\section{Guided Sampling and Policy Refinement}
\label{app:ranger_guided_distillation}

This section explains why guided SDE transitions can refine the unguided flow policy through its transition log-probabilities. We first obtain a Gaussian transition model from flow matching, then show that unguided likelihood fitting of guided samples has a KL-projection interpretation. Finally, we relate this interpretation to the shared flow velocity and the reward-weighted GRPO objective. All distributions below are conditioned on the observation $\mathbf o_k$ unless noted otherwise.

\noindent\phantomsection\label{app:ranger_fm_score}%
\textbf{Flow Matching.}
As in Sec.~\ref{sec:model}, the normalized action sequence $\mathbf a\in\mathbb R^d$, with $d=3H$, is interpolated with $\boldsymbol\epsilon\sim\mathcal N(\mathbf 0,\mathbf I_d)$:
\begin{equation}
    \mathbf x_t=(1-t)\mathbf a+t\boldsymbol\epsilon,
    \qquad t\in[0,1].
    \label{eq:app_ranger_flow_path}
\end{equation}
The sampling direction is $t:1\to0$.
Since $\dot{\mathbf{x}}_t=\boldsymbol{\epsilon}-\mathbf{a}$,
the population minimizer of the flow-matching objective is:
\begin{equation}
    \mathbf v^*(\mathbf x,t,\mathbf o_k)=
    \mathbb E[\boldsymbol\epsilon-\mathbf a\mid
    \mathbf x_t=\mathbf x,\mathbf o_k].
    \label{eq:app_ranger_optimal_velocity}
\end{equation}
Rearranging Eq.~\eqref{eq:app_ranger_flow_path} gives:
\begin{equation}
    \mathbb E[\mathbf a\mid\mathbf x_t=\mathbf x,\mathbf o_k]
    =\mathbf x-t\mathbf v^*(\mathbf x,t,\mathbf o_k).
    \label{eq:app_ranger_clean_estimate}
\end{equation}
Let $p_t(\mathbf x\mid \mathbf o_k)$ be the density of $\mathbf x_t$. The conditional interpolation is Gaussian:
\begin{equation}
    p_t(\mathbf x\mid\mathbf a,\mathbf o_k)
    =\mathcal N(\mathbf x;(1-t)\mathbf a,t^2\mathbf I_d).
    \label{eq:app_ranger_conditional_path}
\end{equation}
For $0<t<1$, the conditional-score identity and Eq.~\eqref{eq:app_ranger_clean_estimate} yield:
\begin{align}
    \nabla_{\mathbf x}\log p_t(\mathbf x\mid \mathbf o_k)
    &=-\frac{\mathbf x-(1-t)\mathbb E[\mathbf a\mid\mathbf x_t=\mathbf x,\mathbf o_k]}{t^2}
      \nonumber\\
    &=-\frac{\mathbf x+(1-t)\mathbf v^*(\mathbf x,t,\mathbf o_k)}{t}.
    \label{eq:app_ranger_score_velocity}
\end{align}
This score--velocity identity follows the time convention of RANGER. The learned velocity $\mathbf v_\theta$ approximates $\mathbf v^*$.

\noindent\phantomsection\label{app:ranger_flow_sde}%
\textbf{Stochastic Flow.}
The ideal flow obeys the continuity equation:
\begin{equation}
    \partial_t p_t=-\nabla_{\mathbf x}\!\cdot(p_t\mathbf v^*).
    \label{eq:app_ranger_continuity}
\end{equation}
Consider the following SDE, integrated backward in $t$:
\begin{equation}
    \mathrm d\mathbf x_t=\mathbf b^*(\mathbf x_t,t,\mathbf o_k)\,\mathrm dt
       +\sigma(t)\,\mathrm d\bar{\mathbf W}_t,
    \label{eq:app_ranger_reverse_sde}
\end{equation}
The corresponding reverse-time Fokker--Planck equation is:
\begin{equation}
    -\partial_t p_t
    =\nabla_{\mathbf x}\!\cdot(p_t\mathbf b^*)
     +\frac{\sigma^2(t)}{2}\nabla_{\mathbf x}^2p_t.
    \label{eq:app_ranger_fp}
\end{equation}
Choose the drift as follows:
\begin{equation}
    \mathbf b^*=\mathbf v^*-\frac{\sigma^2(t)}{2}
                \nabla_{\mathbf x}\log p_t
    \label{eq:app_ranger_score_sde_drift}
\end{equation}
This choice makes the two diffusion terms cancel because $p_t\nabla\log p_t=\nabla p_t$. Thus Eq.~\eqref{eq:app_ranger_fp} reduces to the negative of Eq.~\eqref{eq:app_ranger_continuity}. Combining Eqs.~\eqref{eq:app_ranger_score_velocity} and~\eqref{eq:app_ranger_score_sde_drift}, and replacing $\mathbf v^*$ by $\mathbf v_\theta$, gives the drift used in Sec.~\ref{sec:flow_grpo}:
\begin{equation}
    \mathbf b_\theta=\mathbf v_\theta+
    \frac{\sigma^2(t)}{2t}
    [\mathbf x_t+(1-t)\mathbf v_\theta].
    \label{eq:app_ranger_learned_drift}
\end{equation}
The equality of continuous-time marginals is exact for $\mathbf v^*$; for a learned model and discrete sampling it is approximate. The SDE nevertheless defines the stochastic transitions used to evaluate policy likelihoods.

\noindent\phantomsection\label{app:ranger_guided_gaussians}%
\textbf{Guided Sampling.}
The guidance direction in Eq.~\eqref{eq:guidance_direction} of the main text is $\mathbf u_g(\mathbf x_t,t)=\nabla_{\mathbf x_t}\mathcal E_s(\hat{\boldsymbol\tau})$, where $\hat{\boldsymbol\tau}$ is obtained from the frozen reference flow. In the guided SDE of Eq.~\eqref{eq:guided_flow_sde}, this direction modifies the drift but does not alter the diffusion coefficient.

For the decreasing time grid $1=t_0>\cdots>t_N=0$, let $h_n=t_n-t_{n+1}>0$ and $\sigma_n^2=\sigma^2(t_n)h_n$. Here $\mathbf b_{\theta,n}=\mathbf b_\theta(\mathbf x_n,t_n,\mathbf o_k)$. The unguided Euler--Maruyama transition has mean and density:
\begin{align}
    \boldsymbol\mu_{\theta,n}^0
      &=\mathbf x_n-h_n\mathbf b_{\theta,n},
      \nonumber\\
    p_{\theta,n}^0(\mathbf x_{n+1}\mid\mathbf x_n,\mathbf o_k)
      &=\mathcal N(\mathbf x_{n+1};
         \boldsymbol\mu_{\theta,n}^0,\sigma_n^2\mathbf I_d).
    \label{eq:app_ranger_unguided_density}
\end{align}
The same definitions apply to the old policy. The guidance-induced displacement is $\boldsymbol\Delta_n=-h_n\gamma(t_n)\mathbf u_g(\mathbf x_n,t_n)$, fixed given $\mathbf x_n$, the observation and the training-only guidance context. A guided transition from the old policy is:
\begin{equation}
    \mathbf x_{n+1}=\boldsymbol\mu_{\mathrm{old},n}^0
       +\boldsymbol\Delta_n+\sigma_n\boldsymbol\epsilon_n,
    \quad\boldsymbol\epsilon_n\sim\mathcal N(\mathbf0,\mathbf I_d).
    \label{eq:app_ranger_guided_sample}
\end{equation}
Its conditional behavior distribution is therefore:
\begin{equation}
    q_{\mathrm{old},n}^g
    =\mathcal N(\boldsymbol\mu_{\mathrm{old},n}^0
         +\boldsymbol\Delta_n,\sigma_n^2\mathbf I_d).
    \label{eq:app_ranger_guided_kernel}
\end{equation}
Guidance shifts the conditional mean while preserving the covariance. The argument below depends only on this displacement, not on its particular implementation.

\noindent\phantomsection\label{app:ranger_unguided_likelihood}%
\textbf{Unguided Likelihood.}
Although $\mathbf x_{n+1}$ is sampled from $q_{\mathrm{old},n}^g$, its density under the unguided policy is well defined:
\begin{align}
    &\log p_{\theta,n}^0(\mathbf x_{n+1}\mid\mathbf x_n,\mathbf o_k)
    \nonumber\\
    &\quad=-\frac d2\log(2\pi\sigma_n^2)
      -\frac{\|\mathbf x_{n+1}-\boldsymbol\mu_{\theta,n}^0\|_2^2}
             {2\sigma_n^2}.
    \label{eq:app_ranger_log_density}
\end{align}
As in Eq.~\eqref{eq:unguided_likelihood_ratio} of the main text, the likelihood ratio compares the current and old \emph{unguided} policies. For a fixed observation $\mathbf o_k$ and candidate $j$, we abbreviate $\rho_{k,j,n}$ as $\rho_n(\theta)$:
\begin{equation}
    \rho_n(\theta)=
    \frac{p_{\theta,n}^0(\mathbf x_{n+1}\mid\mathbf x_n,\mathbf o_k)}
         {p_{\mathrm{old},n}^0(\mathbf x_{n+1}\mid\mathbf x_n,\mathbf o_k)}.
    \label{eq:app_ranger_full_ratio}
\end{equation}
Write:
\begin{equation}
    \delta\boldsymbol\mu_n
    =\boldsymbol\mu_{\theta,n}^0
     -\boldsymbol\mu_{\mathrm{old},n}^0.
    \label{eq:app_ranger_mean_delta}
\end{equation}
Substituting the guided sample in Eq.~\eqref{eq:app_ranger_guided_sample} into the two Gaussian log-densities gives:
\begin{align}
    \log\rho_n(\theta)
    &=\frac{\|\boldsymbol\Delta_n+\sigma_n\boldsymbol\epsilon_n\|_2^2
       -\|\boldsymbol\Delta_n+\sigma_n\boldsymbol\epsilon_n
                    -\delta\boldsymbol\mu_n\|_2^2}{2\sigma_n^2}
        \nonumber\\
    &=\frac{(\boldsymbol\Delta_n+\sigma_n\boldsymbol\epsilon_n)^\top
                 \delta\boldsymbol\mu_n}{\sigma_n^2}
       -\frac{\|\delta\boldsymbol\mu_n\|_2^2}{2\sigma_n^2}.
    \label{eq:app_ranger_expanded_ratio}
\end{align}
Thus the unguided log-ratio contains a term aligning the student mean change with $\boldsymbol\Delta_n$, even though neither policy density includes a guidance term.

For comparison, if the same fixed displacement $\boldsymbol\Delta_n$ is inserted into \emph{both} policy means, their guided-to-guided log-ratio instead satisfies:
\begin{equation}
    \mathbb E_{q_{\mathrm{old},n}^g}
    \left[\log\frac{q_{\theta,n}^g}{q_{\mathrm{old},n}^g}\right]
    =-\frac{\|\delta\boldsymbol\mu_n\|_2^2}{2\sigma_n^2}.
    \label{eq:app_ranger_guided_cancel}
\end{equation}
Here $q_{\theta,n}^g$ uses mean $\boldsymbol\mu_{\theta,n}^0+\boldsymbol\Delta_n$. The displacement cancels, so unweighted matching of these guided kernels would not transfer it to the unguided mean.

\noindent\phantomsection\label{app:ranger_kl_projection}%
\textbf{KL Projection.}
Fix the current state, observation and guidance context, and take the expectation of Eq.~\eqref{eq:app_ranger_expanded_ratio} over $\boldsymbol\epsilon_n$. Using $\mathbb E[\boldsymbol\epsilon_n]=\mathbf0$, we obtain:
\begin{align}
    \mathbb E_{q_{\mathrm{old},n}^g}[\log\rho_n(\theta)]
    &=\frac{\boldsymbol\Delta_n^\top\delta\boldsymbol\mu_n}
            {\sigma_n^2}
       -\frac{\|\delta\boldsymbol\mu_n\|_2^2}{2\sigma_n^2}
       \nonumber\\
    &=\frac{\|\boldsymbol\Delta_n\|_2^2
          -\|\delta\boldsymbol\mu_n-\boldsymbol\Delta_n\|_2^2}
            {2\sigma_n^2}.
    \label{eq:app_ranger_expected_ratio}
\end{align}
For an unrestricted student mean, the unweighted pointwise optimum is:
\begin{equation}
    \delta\boldsymbol\mu_n=\boldsymbol\Delta_n,
    \qquad
    \boldsymbol\mu_{\theta,n}^0=
    \boldsymbol\mu_{\mathrm{old},n}^0+\boldsymbol\Delta_n.
    \label{eq:app_ranger_mean_match}
\end{equation}
The same result follows from the KL identity:
\begin{align}
    &\mathbb E_{q_{\mathrm{old},n}^g}\!\left[\log\rho_n(\theta)\right]
    \nonumber\\
    &\quad=D_{\mathrm{KL}}(q_{\mathrm{old},n}^g\|p_{\mathrm{old},n}^0)
       -D_{\mathrm{KL}}(q_{\mathrm{old},n}^g\|p_{\theta,n}^0),
    \label{eq:app_ranger_kl_identity}
\end{align}
where the first term is constant with respect to $\theta$. For the equal-covariance Gaussian kernels, we obtain:
\begin{equation}
    D_{\mathrm{KL}}(q_{\mathrm{old},n}^g\|p_{\theta,n}^0)
    =\frac{\|\boldsymbol\Delta_n-\delta\boldsymbol\mu_n\|_2^2}
           {2\sigma_n^2}.
    \label{eq:app_ranger_gaussian_kl}
\end{equation}
Consequently, maximizing the expected unguided log-ratio on guided samples is equivalent to minimizing the \emph{forward KL} from the guided transition to the unguided student. This equivalence holds for unweighted likelihood fitting. By the chain rule for KL, summing the corresponding conditional terms also gives a path-distribution interpretation for unweighted fitting over the complete SDE sampling chain.

\noindent\phantomsection\label{app:ranger_velocity_refinement}%
\textbf{Flow Refinement.}
The unguided transition mean is determined by the flow velocity. From Eq.~\eqref{eq:app_ranger_unguided_density}, we obtain:
\begin{equation}
    \delta\boldsymbol\mu_n
    =-h_n(\mathbf b_{\theta,n}-\mathbf b_{\mathrm{old},n}).
    \label{eq:app_ranger_drift_mean_relation}
\end{equation}
Here $\mathbf v_{\theta,n}=\mathbf v_\theta(\mathbf x_n,t_n,\mathbf o_k)$, with the same shorthand for the old policy. Using Eq.~\eqref{eq:app_ranger_learned_drift}, the state-dependent term common to both drifts cancels, giving:
\begin{align}
    \delta\boldsymbol\mu_n
    =-h_n\!\left(1+\frac{(1-t_n)\sigma^2(t_n)}{2t_n}\right)
    (\mathbf v_{\theta,n}-\mathbf v_{\mathrm{old},n}).
    \label{eq:app_ranger_velocity_shift}
\end{align}
Combining this relation with the ideal mean-matching condition in Eq.~\eqref{eq:app_ranger_mean_match} yields:
\begin{equation}
    \mathbf v_{\theta,n}-\mathbf v_{\mathrm{old},n}
    =-\frac{\boldsymbol\Delta_n}{h_n}
     \left(1+\frac{(1-t_n)\sigma^2(t_n)}{2t_n}\right)^{-1}.
    \label{eq:app_ranger_velocity_distill}
\end{equation}
This is the local optimum of unweighted likelihood matching, not an exact update rule for a finite-capacity network. It shows how fitting guided transitions can change the \emph{unguided} velocity field used at inference.

\noindent\phantomsection\label{app:ranger_reward_selective}%
\textbf{GRPO Refinement.}
Rather than imitating all guided samples, Ranger-GRPO
selectively reinforces promising trajectories using
skill-specific rewards and group-relative advantages.
For each guided transition, the gradient of the
clipped reward surrogate at $\theta=\theta_{\mathrm{old}}$, where the ratio is one, is:
\begin{align}
&A_{k,j}\nabla_\theta
\log p_{\theta,n}^0(\mathbf x_{n+1}\mid\mathbf x_n,\mathbf o_k)
\big|_{\theta=\theta_{\mathrm{old}}}
\nonumber\\
&\quad=
\frac{A_{k,j}}{\sigma_n^2}
(\boldsymbol\Delta_n+\sigma_n\boldsymbol\epsilon_n)^\top
\nabla_\theta\boldsymbol\mu_{\theta,n}^0
\big|_{\theta=\theta_{\mathrm{old}}}.
\label{eq:app_ranger_adv_gradient}
\end{align}
This gradient increases the unguided likelihood
of positive-advantage guided transitions while
suppressing negative-advantage ones.
Together with the KL-projection interpretation,
it explains how Ranger-GRPO transfers
high-reward guided behaviors into the shared
flow velocity field $\mathbf v_\theta$.
Consequently, inference requires only the
unguided ODE
$\mathrm d\mathbf x_t=\mathbf v_\theta(\mathbf x_t,t,\mathbf o_k)\,\mathrm dt$,
without skill labels, privileged information,
or guidance.

\section{Ranger-GRPO Training Procedure}
\label{sec:appendix_algorithm}

The online sampling and shared-policy refinement
procedure is summarized in Algorithm~\ref{alg:ranger_grpo}.

\noindent\phantomsection\label{app:contents_training_hyperparameters}%
\textbf{Training Hyperparameters.}
For each observation, we sample $G=16$ candidate
trajectories using guided SDE sampling with $N=5$
discretization steps.
We train for $I=400$ rounds. Each round uses two minibatches,
with one AdamW update per minibatch (800 updates in total).
The learning rate and weight decay are
$3\times10^{-4}$ and $10^{-4}$, respectively.
We use a likelihood-ratio clipping parameter
$\varepsilon_{\mathrm{clip}}=10^{-4}$,
a KL coefficient $\beta_{\mathrm{KL}}=0.04$,
and a gradient norm clipping threshold of $1.0$.
Post-training is performed on a single
NVIDIA GeForce RTX 4090 GPU.

\noindent\phantomsection\label{app:contents_noise_schedule}%
\textbf{Noise Schedule.}
We use a uniform reverse-time grid
$t_n=1-n/5$ for $n=0,\ldots,5$,
with step size $h_n=0.2$ and noise level $\eta=0.9$.
The diffusion coefficient is defined as:
\begin{equation}
\sigma(t)=\eta\sqrt{\frac{t}{\max(1-t,\,0.2)}},
\qquad t\in[0,1].
\end{equation}
The denominator is bounded below by $0.2$ to avoid
the singularity at $t=1$.
The discrete transition standard deviation is
$\sigma_n=\sigma(t_n)\sqrt{h_n}$.
All five transitions remain stochastic and are used
for policy optimization.

\noindent\phantomsection\label{app:contents_guidance_implementation}%
\textbf{Guidance Implementation.}
At each guided step, the clean action is estimated
by integrating the frozen SFT flow from the current
sampling state toward $t=0$.
The skill-energy gradient is backpropagated through
the reference integration to the current sampling state,
while the SFT parameters remain frozen.
Guidance is applied only to the final two of the five
SDE steps. The gradient is normalized by its
per-trajectory RMS magnitude, and the guidance
displacement is scaled by the discrete transition
noise standard deviation.
We use an initial guidance strength $\alpha_0=0.2$,
which follows a cosine annealing schedule:
\begin{equation}
\alpha_r=
\frac{\alpha_0}{2}
\left[
1+\cos\left(\frac{\pi r}{I}\right)
\right],
\label{eq:guidance_annealing}
\end{equation}
where $r=1,\ldots,I$ denotes the training round.

\begin{algorithm}[t]
\caption{Ranger-GRPO: Guided Multi-Skill Policy Refinement}
\label{alg:ranger_grpo}
\small
\begin{algorithmic}[1]
\setlength{\itemsep}{0pt}
\setlength{\parskip}{0pt}
\setlength{\parsep}{0pt}
\setlength{\topsep}{0pt}
\setlength{\partopsep}{0pt}
\linespread{0.94}\selectfont

\Require Simulator, skill set $\mathcal S$,
SFT policy $\theta_{\mathrm{SFT}}$,
group size $G$, sampling steps $N$, training rounds $I$

\State Initialize $\theta\leftarrow\theta_{\mathrm{SFT}}$
\State Freeze $\theta_{\mathrm{ref}}\leftarrow\theta_{\mathrm{SFT}}$
\State Use $\mathbf x_n:=\mathbf x_{t_n}$ on the grid $1=t_0>\cdots>t_N=0$

\For{$r=1,\ldots,I$}
    \State $\theta_{\mathrm{old}}\leftarrow\theta$;
    $\mathcal U\leftarrow\emptyset$
    \State Update guidance schedule
    \State Collect skill-balanced observations
    $\mathcal B=\{(s,\mathbf o_k)\}$
    \\
    \Statex \textit{// Guided SDE sampling}
    \ForAll{$(s,\mathbf o_k)\in\mathcal B$}
        \For{$j=1,\ldots,G$}
            \State Sample
            $\mathbf x_{t_0}\sim\mathcal N(\mathbf0,\mathbf I_d)$

            \For{$n=0,\ldots,N-1$}
                \If{guidance is enabled at step $n$}
                    \State $\mathbf x_n^{\mathrm{ref}}
                    \leftarrow\mathbf x_n$

                    \For{$m=n,\ldots,N-1$}
                        \State $\mathbf x_{m+1}^{\mathrm{ref}}
                        \leftarrow\mathbf x_m^{\mathrm{ref}}
                        -h_m\mathbf v_{\mathrm{ref}}
                        (\mathbf x_m^{\mathrm{ref}},t_m,\mathbf o_k)$
                    \EndFor

                    \State $\hat{\boldsymbol\tau}
                    \leftarrow\mathcal T(\mathbf x_N^{\mathrm{ref}})$

                    \State $\mathbf u_{g,n}\leftarrow
                    \nabla_{\mathbf x_n}
                    \mathcal E_s(\hat{\boldsymbol\tau})$

                    \State Compute $\boldsymbol\Delta_n$
                \Else
                    \State $\boldsymbol\Delta_n\leftarrow\mathbf0$
                \EndIf

                \State Sample $\mathbf x_{n+1}
                \sim q^g_{\mathrm{old},n}$
                using Eq.~\eqref{eq:guided_gaussian_policy}

                \State Store transition
                $(k,j,n,\mathbf o_k,\mathbf x_n,\mathbf x_{n+1})$
                in $\mathcal U$
            \EndFor

            \State $\boldsymbol\tau_{k,j}
            \leftarrow\mathcal T(\mathbf x_N)$
            \State $R_{k,j}\leftarrow
            R_s(\boldsymbol\tau_{k,j})$
        \EndFor

        \State Compute group-relative advantages
        $\{A_{k,j}\}_{j=1}^{G}$
        \State Assign each advantage to all transitions
    \EndFor
    \\
    \Statex \textit{// Unguided policy refinement}
    \ForAll{minibatches in $\mathcal U$}
        \State Compute $\rho_{k,j,n}$ using
        Eq.~\eqref{eq:unguided_likelihood_ratio}
        \State Update $\theta$ by maximizing
        Eq.~\eqref{eq:guided_grpo_objective}
    \EndFor
\EndFor

\State \Return refined unguided policy $\theta$
\end{algorithmic}
\end{algorithm}

\section{Skill-Specific Rewards and Guidance Energies}
\label{app:skill_rewards}
\label{app:skill_guidance}

Each skill uses a trajectory reward $R_s$ for advantage
estimation and a differentiable energy $\mathcal E_s$
for sampling guidance. 
During post-training, reward evaluation and guidance use privileged simulator geometry and future pedestrian trajectories. This information is not provided to the deployed policy.
Let $\mathbf p_h=(x_h,y_h)$
denote the robot waypoint at prediction step $h$,
with $H=24$, $\mathbf p_0=\mathbf 0$, and a physical time step
of $\Delta T=0.16\,\mathrm{s}$.
The robot velocity and speed are
$\mathbf v_h=(\mathbf p_h-\mathbf p_{h-1})/\Delta T$
and $s_h=\|\mathbf v_h\|$, respectively.
The superscript $\mathrm{ref}$ denotes a reference quantity,
and $\|\cdot\|$ denotes the Euclidean norm.

\noindent\phantomsection\label{app:common_reward}%
\textbf{Shared Motion and Safety Reward.}
All skills share a reward that balances
navigation progress and collision safety:
\begin{equation}
\begin{aligned}
R_{\mathrm{common}}={}&
-\lambda_{\mathrm{length}}C_{\mathrm{length}}
-\lambda_{\mathrm{progress}}C_{\mathrm{progress}}\\
&-\lambda_{\mathrm{safety}}
(C_{\mathrm{static}}+C_{\mathrm{pedestrian}}).
\end{aligned}
\label{eq:app_shared_reward}
\end{equation}
We set $\lambda_{\mathrm{length}}=1.6$,
$\lambda_{\mathrm{progress}}=0.4$, and
$\lambda_{\mathrm{safety}}=2$.
All four costs are bounded within $[0,1]$.

Trajectory length and goal progress are computed as
$L=\sum_{h=1}^{H}\|\mathbf p_h-\mathbf p_{h-1}\|_2$
and $P=\|\mathbf g\|_2-\|\mathbf g-\mathbf p_H\|_2$,
respectively, where $\mathbf g$ denotes the relative goal.
Their reference values are obtained from the medians
of 16 fixed-noise trajectories generated by the frozen
SFT policy. VO risk adjusts these targets based on
the preferred speed and direction.
$C_{\mathrm{length}}$ and $C_{\mathrm{progress}}$
penalize trajectories falling short of their
respective targets, without penalizing those
that exceed them. In the absence of VO risk,
the original SFT reference values are preserved.

Static-obstacle clearance is evaluated using
the Euclidean distance transform of the occupancy
map, accounting for the robot radius.
The minimum clearance is computed over all
map cells traversed by the trajectory segments,
rather than only the predicted waypoints.
Out-of-map regions are treated as obstacles.
$C_{\mathrm{static}}$ increases linearly as the
minimum clearance falls below
$d_{\mathrm{static}}=0.7\,\mathrm{m}$,
reaching its maximum at zero clearance.

Pedestrian risk is evaluated using time-aligned
robot and pedestrian trajectories, linearly
interpolated between consecutive waypoints.
Let $d_{\mathrm{min}}$ denote the minimum
center-to-center distance over all pedestrians
and trajectory segments.
The pedestrian cost combines proximity risk
with a temporal weight emphasizing earlier encounters:
\begin{equation}
C_{\mathrm{pedestrian}}=
\begin{cases}
1, & d_{\mathrm{min}}<d_{\mathrm{col}},\\
s_{\mathrm{risk}}w_t, & \text{otherwise},
\end{cases}
\label{eq:app_pedestrian_cost}
\end{equation}
The risk score and temporal weight are defined as follows:
\begin{equation}
\begin{aligned}
s_{\mathrm{risk}}
&=\operatorname{clip}\left(
\frac{d_{\mathrm{safe}}-d_{\mathrm{min}}}
{d_{\mathrm{safe}}-d_{\mathrm{col}}},0,1
\right),\\
w_t
&=\frac{1}{2}\left[
1+\operatorname{clip}\left(
1-\frac{t_{\mathrm{min}}}{H\Delta T},0,1
\right)\right].
\end{aligned}
\end{equation}
Here, $d_{\mathrm{col}}=0.45\,\mathrm{m}$
is the collision threshold and
$d_{\mathrm{safe}}=1.2\,\mathrm{m}$
is the pedestrian clearance threshold.
The time $t_{\mathrm{min}}$ denotes the
closest-approach time estimated from
discrete synchronized waypoints, while
clearance is evaluated continuously along
each segment.
Distances below $d_{\mathrm{col}}$ receive
the maximum penalty, whereas distances
beyond $d_{\mathrm{safe}}$ incur no pedestrian cost.

\begin{figure*}[t]
    \centering
    \includegraphics[width=\textwidth]{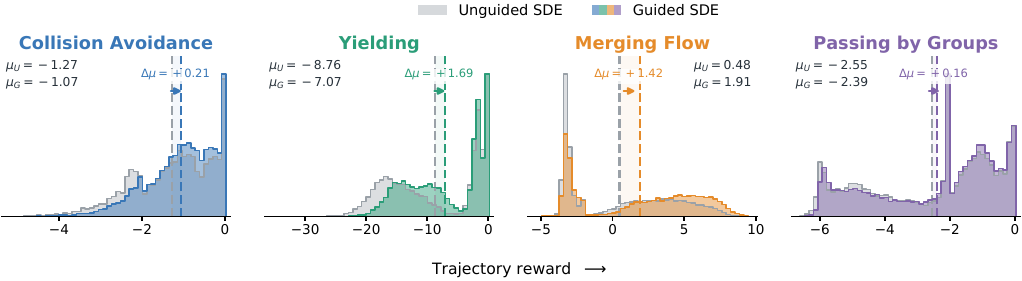}
    \caption{Reward distributions under matched unguided
    and guided SDE sampling across four interaction skills.
    Color-matched dashed lines indicate the corresponding
    distribution means.}
    \label{fig:guided_exploration_analysis}
\end{figure*}

\begin{figure*}[t]
    \centering
    \includegraphics[width=\textwidth]{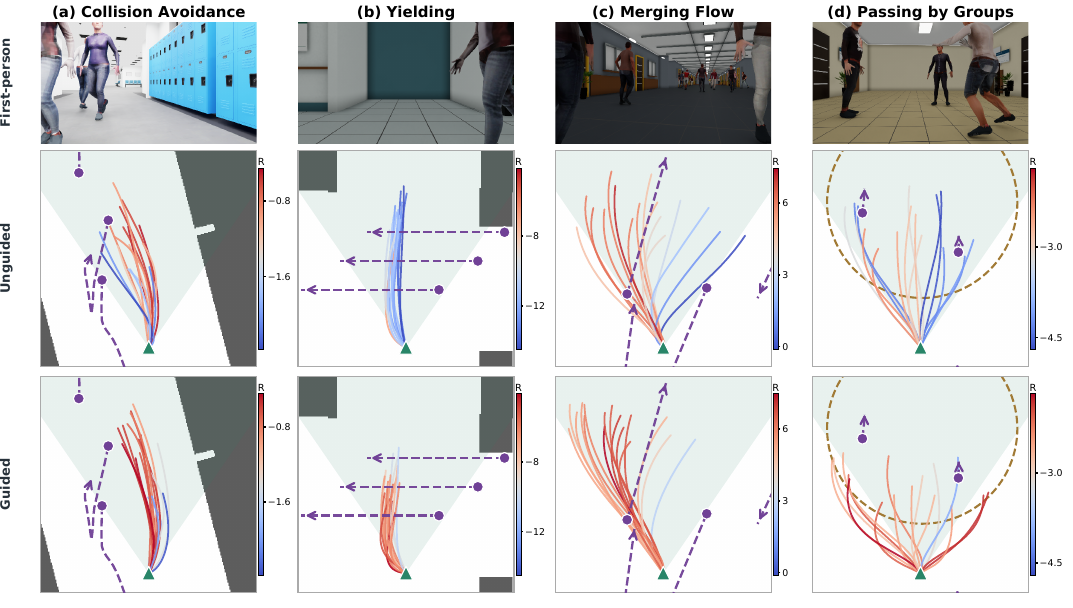}
    \caption{Qualitative effects of skill-specific guidance. From top
    to bottom, the rows show first-person observations, unguided ODE
    candidates, and guided ODE candidates for the four interaction
    skills. Candidate trajectories are colored by trajectory reward.}
    \label{fig:guidance_ode_qualitative}
\end{figure*}

\noindent\phantomsection\label{app:headon_reward}%
\textbf{Collision Avoidance.}
Collision Avoidance combines the shared reward with
a velocity-obstacle (VO) motion preference.
The reward and guidance energy are defined as:
\begin{equation}
\begin{aligned}
R_{\mathrm{avoid}}
&=R_{\mathrm{common}}
-\lambda_{\mathrm{VO}}C_{\mathrm{VO}},\\
\mathcal E_{\mathrm{avoid}}
&=\max(d_{\mathrm{safe}}-d_{\mathrm{min}},0).
\end{aligned}
\label{eq:app_avoid_reward}
\end{equation}
Here, $C_{\mathrm{VO}}$ measures the risk-weighted
deviation of trajectory speed and heading from a
selected VO reference. The reference is selected
from 244 candidate velocities based on collision
feasibility, heading deviation, and speed preservation.
The guidance energy penalizes insufficient
time-aligned robot--pedestrian separation.
We set $\lambda_{\mathrm{VO}}=0.7$ and
$d_{\mathrm{safe}}=1.2\,\mathrm{m}$.
Guidance is disabled when no valid pedestrian is present.

\noindent\phantomsection\label{app:yield_reward}%
\textbf{Yielding.}
Yielding encourages the robot to slow down for
crossing pedestrians and resume motion afterward.
A scripted waiting-and-release reference provides
the longitudinal positions $x_h^{\mathrm{ref}}$
and speeds $s_h^{\mathrm{ref}}$.
The reward and guidance energy are defined as:
\begin{equation}
\begin{aligned}
R_{\mathrm{yield}}
={}&R_{\mathrm{common}}
-C_{\mathrm{lat}}
-\lambda_{\mathrm{time}}C_{\mathrm{time}}\\
&-\lambda_{\mathrm{wait}}C_{\mathrm{wait}}
-\lambda_{\mathrm{resume}}C_{\mathrm{resume}},\\
\mathcal E_{\mathrm{yield}}
={}&\frac{1}{H}\sum_{h=1}^{H}
\Big[
(s_h-s_h^{\mathrm{ref}})^2
+\lambda_{\mathrm{lateral}}y_h^2
\Big].
\end{aligned}
\label{eq:app_yield_reward}
\end{equation}
Here, $C_{\mathrm{lat}}$ penalizes mean squared
lateral deviation, while $C_{\mathrm{time}}$
measures longitudinal reference error using
Smooth-$L_1$ loss.
$C_{\mathrm{wait}}$ penalizes motion during
waiting, whereas $C_{\mathrm{resume}}$ penalizes
speed mismatch during recovery.
Their normalized temporal weights are:
\begin{equation}
\begin{aligned}
w_h^{\mathrm{wait}}
&=\exp\left[
-\left(\frac{s_h^{\mathrm{ref}}}
{v_{\mathrm{scale}}}\right)^2
\right],\\
w_h^{\mathrm{resume}}
&=1-w_h^{\mathrm{wait}}.
\end{aligned}
\end{equation}
We set $\lambda_{\mathrm{time}}=2$,
$\lambda_{\mathrm{wait}}=12$,
$\lambda_{\mathrm{resume}}=6$,
$\lambda_{\mathrm{lateral}}=0.25$, and
$v_{\mathrm{scale}}=0.2\,\mathrm{m/s}$.
The reference speed is clipped to
$[0,1]\,\mathrm{m/s}$.
Guidance is disabled when yielding is not required.

\noindent\phantomsection\label{app:follow_reward}%
\textbf{Merging Flow.}
Merging Flow encourages the robot to align its
motion with nearby same-direction pedestrians
while avoiding opposing flows.
The reward and guidance energy are defined as:
\begin{equation}
\begin{aligned}
R_{\mathrm{merge}}
={}&R_{\mathrm{common}}
+\lambda_{\mathrm{same}}R_{\mathrm{same}}
+R_{\mathrm{speed}}\\
&-\lambda_{\mathrm{opp}}C_{\mathrm{opposite}}
+\lambda_{\mathrm{ref}}R_{\mathrm{ref}},\\
\mathcal E_{\mathrm{merge}}
={}&\frac{\lambda_{\mathrm{track}}}{H}
\sum_{h=1}^{H}
\Big[
\|\mathbf p_h-\mathbf p_h^{\mathrm{ref}}\|_2^2\\
&\qquad+
\lambda_{\mathrm{vel}}
\|\mathbf v_h-\mathbf v_h^{\mathrm{ref}}\|_2^2
\Big].
\end{aligned}
\label{eq:app_follow_reward}
\end{equation}
Here, $R_{\mathrm{same}}$ rewards proximity to
same-direction pedestrian flow,
$R_{\mathrm{speed}}$ encourages velocity matching,
and $C_{\mathrm{opposite}}$ penalizes proximity
to opposing pedestrians. These terms account
for local spatial relationships and pedestrian
motion directions.
$R_{\mathrm{ref}}$ is an exponential imitation
reward based on position and velocity errors.
The reference trajectory is generated by smoothly
merging the robot into nearby same-direction
pedestrian flow while maintaining a
$1.2\,\mathrm{m}$ longitudinal headway.
Among the collision-free candidates, we select
the one that best aligns with the local pedestrian
motion. If no feasible reference is available,
guidance follows the mean same-direction flow
velocity, and $R_{\mathrm{ref}}$ is set to zero.
The guidance energy tracks the reference
positions $\mathbf p_h^{\mathrm{ref}}$ and
velocities $\mathbf v_h^{\mathrm{ref}}$.
We set $\lambda_{\mathrm{same}}=3$,
$\lambda_{\mathrm{opp}}=2$,
$\lambda_{\mathrm{ref}}=10$, and
$\lambda_{\mathrm{track}}=
\lambda_{\mathrm{vel}}=0.5$.

\noindent\phantomsection\label{app:conversation_reward}%
\textbf{Passing by Groups.}
Passing by Groups encourages the robot to navigate
around conversational groups while preserving goal progress.
Each valid group $i$ is represented by its center
$\mathbf c_i$ and enclosing radius $r_i$.
For each predicted waypoint, we define the normalized
social-space intrusion as:
\begin{equation}
z_h=\max_i
\frac{r_i+m_{\mathrm{group}}
-\|\mathbf p_h-\mathbf c_i\|_2}
{\delta_{\mathrm{group}}},
\end{equation}
where $m_{\mathrm{group}}=0.65\,\mathrm{m}$ is the
social clearance margin and
$\delta_{\mathrm{group}}=0.2\,\mathrm{m}$
controls the transition sharpness.
The reward and guidance energy are defined as:
\begin{equation}
\begin{aligned}
R_{\mathrm{group}}
={}&R_{\mathrm{common}}
-\frac{4}{H}\sum_{h=1}^{H}
\operatorname{sigmoid}(z_h),\\
\mathcal E_{\mathrm{group}}
={}&\lambda_{\mathrm{group}}
\Bigg[
\frac{1}{H}\sum_{h=1}^{H}
\operatorname{softplus}(z_h)\\
&\quad+
\left[
\|\mathbf p_H-\mathbf g\|_2
-\|\mathbf p_H^{\mathrm{ref}}-\mathbf g\|_2
\right]_+^2
\Bigg],
\end{aligned}
\label{eq:app_group_reward}
\end{equation}
where $\lambda_{\mathrm{group}}=0.2$ and
$[z]_+=\max(z,0)$.
The bounded sigmoid penalty evaluates social-space
intrusion, while the softplus energy maintains
effective guidance gradients for severe intrusion.
The terminal term penalizes reduced goal progress
relative to the stored SFT reference trajectory.
Only groups with at least two visible members are
considered. When no valid group is present,
both the group-specific reward penalty and
guidance energy are set to zero.

\begin{figure*}[t]
    \centering
    \includegraphics[width=\textwidth]{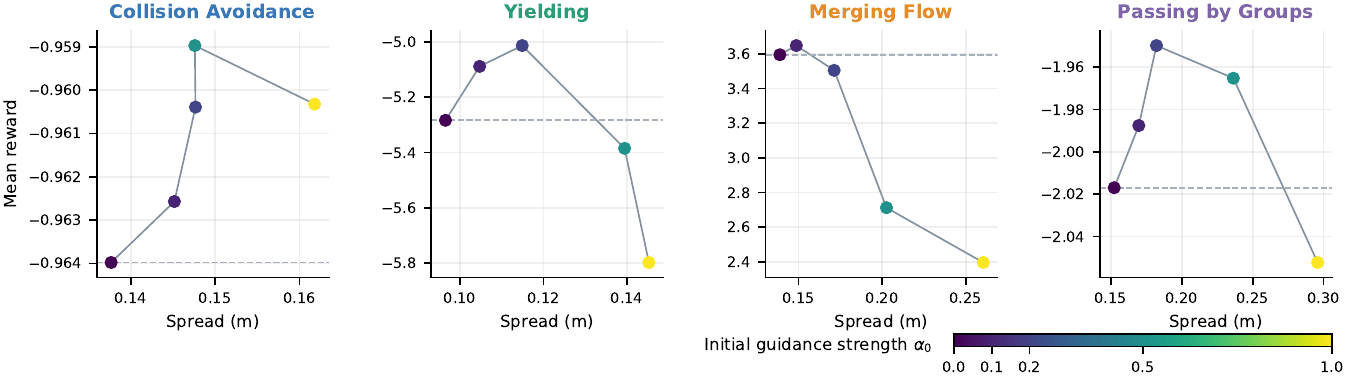}
    \caption{Mean trajectory reward versus spread for five
    initial guidance strengths $\alpha_0$.
    Dashed horizontal lines indicate the mean reward
    of the Flow-GRPO baseline.}
    \label{fig:guidance_reward_spread_ablation}
\end{figure*}

\begin{table*}[t]
\centering
\footnotesize
\setlength{\tabcolsep}{2.0pt}
\begin{minipage}[t]{0.238\textwidth}
\vspace{0pt}
\centering
\textit{Collision Avoidance}\par
\vspace{1mm}
\noindent
\resizebox{\linewidth}{!}{%
\begin{tabular}[t]{@{}l@{\hspace{1pt}}rrrrr@{}}
\toprule
$\alpha_0$ & Mean $\uparrow$ & Top-1 $\uparrow$ & P-Col $\downarrow$
& Prog. $\uparrow$ & Spread $\uparrow$ \\
\midrule
SFT & -1.121 & -0.749 & 13.26 & 2.883 & \textbf{0.163} \\
$0.0_{\scriptscriptstyle\text{(Flow-GRPO)}}$ & -0.964 & -0.714 & \textbf{11.76} & 2.937 & 0.138 \\
0.1 & -0.963 & -0.690 & 12.03 & 2.993 & 0.145 \\
0.2 & -0.960 & -0.685 & 12.16 & 2.997 & 0.148 \\
0.5 & \textbf{-0.959} & -0.691 & 13.57 & 3.039 & 0.148 \\
1.0 & -0.960 & \textbf{-0.669} & 13.03 & \textbf{3.055} & 0.162 \\
\bottomrule
\end{tabular}
}
\end{minipage}\hfill
\begin{minipage}[t]{0.238\textwidth}
\vspace{0pt}
\centering
\emph{Yielding}\\[1mm]
\resizebox{\linewidth}{!}{%
\begin{tabular}[t]{@{}l@{\hspace{1pt}}rrrrr@{}}
\toprule
$\alpha_0$ & Mean $\uparrow$ & Top-1 $\uparrow$ & P-Col $\downarrow$
& Prog. $\uparrow$ & Spread $\uparrow$ \\
\midrule
SFT & -8.599 & -7.357 & 17.65 & \textbf{3.355} & 0.098 \\
$0.0_{\scriptscriptstyle\text{(Flow-GRPO)}}$ & -5.284 & -4.596 & 12.11 & 2.798 & 0.097 \\
0.1 & -5.089 & -4.359 & \textbf{11.88} & 2.711 & 0.105 \\
0.2 & \textbf{-5.014} & \textbf{-4.179} & 12.07 & 2.657 & 0.115 \\
0.5 & -5.385 & -4.402 & 12.99 & 2.667 & 0.139 \\
1.0 & -5.797 & -4.701 & 13.40 & 2.723 & \textbf{0.145} \\
\bottomrule
\end{tabular}
}
\end{minipage}\hfill
\begin{minipage}[t]{0.238\textwidth}
\vspace{0pt}
\centering
\emph{Merging Flow}\\[1mm]
\resizebox{\linewidth}{!}{%
\begin{tabular}[t]{@{}l@{\hspace{1pt}}rrrrr@{}}
\toprule
$\alpha_0$ & Mean $\uparrow$ & Top-1 $\uparrow$ & P-Col $\downarrow$
& Prog. $\uparrow$ & Spread $\uparrow$ \\
\midrule
SFT & 1.178 & 2.858 & 39.80 & 2.144 & 0.163 \\
$0.0_{\scriptscriptstyle\text{(Flow-GRPO)}}$ & 3.596 & 4.819 & 22.79 & 2.336 & 0.139 \\
0.1 & \textbf{3.649} & 4.899 & \textbf{22.64} & 2.329 & 0.149 \\
0.2 & 3.506 & \textbf{4.906} & 23.59 & 2.415 & 0.171 \\
0.5 & 2.712 & 4.427 & 29.38 & \textbf{2.424} & 0.203 \\
1.0 & 2.396 & 4.785 & 32.44 & 2.382 & \textbf{0.260} \\
\bottomrule
\end{tabular}
}
\end{minipage}\hfill
\begin{minipage}[t]{0.238\textwidth}
\vspace{0pt}
\centering
\emph{Passing by Groups}\\[1mm]
\resizebox{\linewidth}{!}{%
\begin{tabular}[t]{@{}l@{\hspace{1pt}}rrrrr@{}}
\toprule
$\alpha_0$ & Mean $\uparrow$ & Top-1 $\uparrow$ & P-Col $\downarrow$
& Prog. $\uparrow$ & Spread $\uparrow$ \\
\midrule
SFT & -2.567 & -2.220 & 27.46 & \textbf{2.425} & 0.157 \\
$0.0_{\scriptscriptstyle\text{(Flow-GRPO)}}$ & -2.017 & -1.715 & 11.05 & 2.108 & 0.152 \\
0.1 & -1.988 & -1.661 & 11.07 & 2.093 & 0.170 \\
0.2 & \textbf{-1.950} & -1.599 & \textbf{10.60} & 2.102 & 0.182 \\
0.5 & -1.965 & \textbf{-1.542} & 12.08 & 2.115 & 0.236 \\
1.0 & -2.052 & -1.556 & 14.68 & 2.286 & \textbf{0.296} \\
\bottomrule
\end{tabular}
}
\end{minipage}
\caption{Guidance-strength ablation after 100 training rounds.
The $\alpha_0=0$ (Flow-GRPO) setting follows the same
single-skill training protocol with guidance disabled.
Bold indicates the best result for each metric within
each skill.}
\label{tab:guidance_strength}
\end{table*}

\section{Effects of Skill-Specific Guidance}
\label{app:guidance_qualitative}

\noindent\phantomsection\label{app:contents_guided_sde}%
\textbf{Guided SDE Exploration.}
To assess whether skill-energy guidance steers exploration toward high-reward behaviors, we compare guided and unguided sampling on the open-loop evaluation set. For the quantitative analysis, both samplers use
the same frozen SFT policy, observations, initial latents, and
stochastic noise to generate 16 trajectories per observation.
We consider observations for which the corresponding
skill guidance is applicable.
As shown in Fig.~\ref{fig:guided_exploration_analysis}, guided SDE
sampling shifts the reward distributions toward higher values for all
four skills, improving the mean rewards from $-1.27$ to $-1.07$,
$-8.76$ to $-7.07$, $0.48$ to $1.91$, and $-2.55$ to $-2.39$,
respectively.

\noindent\phantomsection\label{app:contents_trajectory_generation}%
\textbf{Guidance Effects on Trajectory Generation.}
While the SDE comparison measures the exploration distribution used
during training, an ODE comparison provides a clearer qualitative view
of how guidance changes trajectory geometry without additional
transition noise. We therefore visualize matched guided and unguided
ODE samples using the same SFT checkpoint and identical initial
latents. Fig.~\ref{fig:guidance_ode_qualitative} uses the simplified
deterministic update:
\begin{equation}
\mathbf x_{n+1}^{g}
=\mathbf x_n^{g}-h_n\mathbf v_{\mathrm{ref}}
(\mathbf x_n^{g},t_n,\mathbf o_k)
+\boldsymbol\Delta_n^{g}.
\end{equation}
At guided steps, we compute $\boldsymbol\Delta_n^g$ by differentiating
the skill energy through the frozen SFT Euler rollout from
$\mathbf x_n^g$ to $t=0$.
At other steps, $\boldsymbol\Delta_n^g=\mathbf 0$.
Both analyses use five sampling steps, with guidance applied only in the
final two steps at a fixed
strength of $\alpha=0.2$.
As illustrated in Fig.~\ref{fig:guidance_ode_qualitative},
guidance produces higher-reward trajectories across all four scenarios,
encouraging safer pedestrian avoidance, yielding to crossing
pedestrians, alignment with pedestrian flows, and detouring around
conversational groups.
Together, these results show that skill-energy guidance
steers exploration toward desirable interaction behaviors that are less
readily sampled by the unguided policy, providing higher-quality
candidates for reinforcement post-training.

\noindent\phantomsection\label{app:contents_guidance_ablation}%
\textbf{Guidance-Strength Ablation.}
To examine sensitivity to the guidance strength in post-training, we vary the initial coefficient $\alpha_0$ while
keeping the SFT initialization, rewards, observation order, noise, and
optimizer settings fixed. These are independent single-skill
sensitivity runs, each trained for 100 rounds with 64 observations per
round, yielding 6,400 observation draws and 102,400 candidate
trajectories. The selected $\alpha_0$ is annealed to zero over the
100 rounds. We report the mean over five unguided ODE evaluations.
Each repeat uses a different set of initial noise samples, shared across methods.
We use the open-loop metrics defined in the main text.
Table~\ref{tab:guidance_strength} shows that stronger guidance
generally increases trajectory diversity, while excessive
guidance can degrade mean reward and collision safety.
The optimal strength varies across skills:
$\alpha_0=0.2$ achieves the highest mean reward for
\emph{Yielding} and \emph{Passing by Groups},
whereas \emph{Merging Flow} favors $\alpha_0=0.1$.
The mean reward of \emph{Collision Avoidance} remains
relatively insensitive to guidance strength.
Notably, $\alpha_0=0.2$ consistently improves both Top-1
reward and trajectory spread over unguided Flow-GRPO
across all four skills.
Fig.~\ref{fig:guidance_reward_spread_ablation} visualizes
the corresponding reward--diversity trade-off.
Based on these observations, we adopt a shared initial
guidance strength of $\alpha_0=0.2$ for four-skill
joint post-training.

\section{Skill-Specific Closed-Loop Analysis}
\label{app:qualitative_diagnostics}
While overall navigation metrics evaluate task success
and collision safety, they do not fully characterize
the interaction behaviors acquired through post-training.
We therefore conduct skill-specific closed-loop analyses
to examine how RANGER adapts its spatial and temporal
responses to different social interactions, combining
quantitative behavioral metrics with qualitative
trajectory and motion-profile comparisons.

\noindent\phantomsection\label{app:contents_quantitative}%
\textbf{Quantitative Analysis.}
We evaluate four social navigation skills over
100 matched closed-loop episodes per method and skill.
All metrics are computed from executed trajectories
over the corresponding interaction intervals within
each episode and then averaged across episodes.

For \emph{Collision Avoidance}, we measure the minimum
distance $d_{\mathrm{CA}}$ between the robot and
oncoming pedestrians during active encounters.
For \emph{Yielding}, we measure the speed reduction
$\Delta v_{\mathrm Y}$ and the minimum crossing
clearance $d_{\mathrm Y}$. The speed reduction is
computed as the difference between the mean robot
speeds before and during pedestrian crossing,
clipped at zero. The crossing clearance is the
minimum robot--pedestrian distance during the
crossing interval.
For \emph{Merging Flow}, we measure the mean absolute
lateral position error $e_{\mathrm{MF}}^p$ and the
mean Euclidean velocity error $e_{\mathrm{MF}}^v$
relative to the surrounding same-direction pedestrian
flow. The reference lateral position is the midpoint
of the lateral pedestrian-flow span, and the reference
velocity is the mean velocity of same-direction
pedestrians. Both errors are computed over frames
with at least two such pedestrians in the active
corridor.
For \emph{Passing by Groups}, we measure the minimum
distance $d_{\mathrm{PG}}$ between the robot and
valid conversational-group centers during active
group-passing interactions.

As shown in Table~\ref{tab:skill_closedloop_metrics},
RANGER achieves the best results on four of the six
skill-specific metrics.
In \emph{Collision Avoidance}, it maintains the
largest pedestrian clearance ($0.929\,\mathrm{m}$).
In \emph{Yielding}, RANGER exhibits the strongest
speed reduction ($0.467\,\mathrm{m/s}$) while
preserving the crossing clearance ($1.395\,\mathrm{m}$).
For \emph{Merging Flow}, RANGER achieves the
lowest lateral alignment error ($0.911\,\mathrm{m}$),
although its velocity error remains higher than
those of X-NavDP and Flow-GRPO.
In \emph{Passing by Groups}, RANGER maintains
the greatest separation from conversational group centers
($2.572\,\mathrm{m}$).
Together, these results support RANGER's ability
to acquire distinct interaction behaviors,
including pedestrian avoidance, yielding,
flow alignment, and socially aware group passing.

\begin{figure*}[!t]
    \centering
    \includegraphics[width=\textwidth]{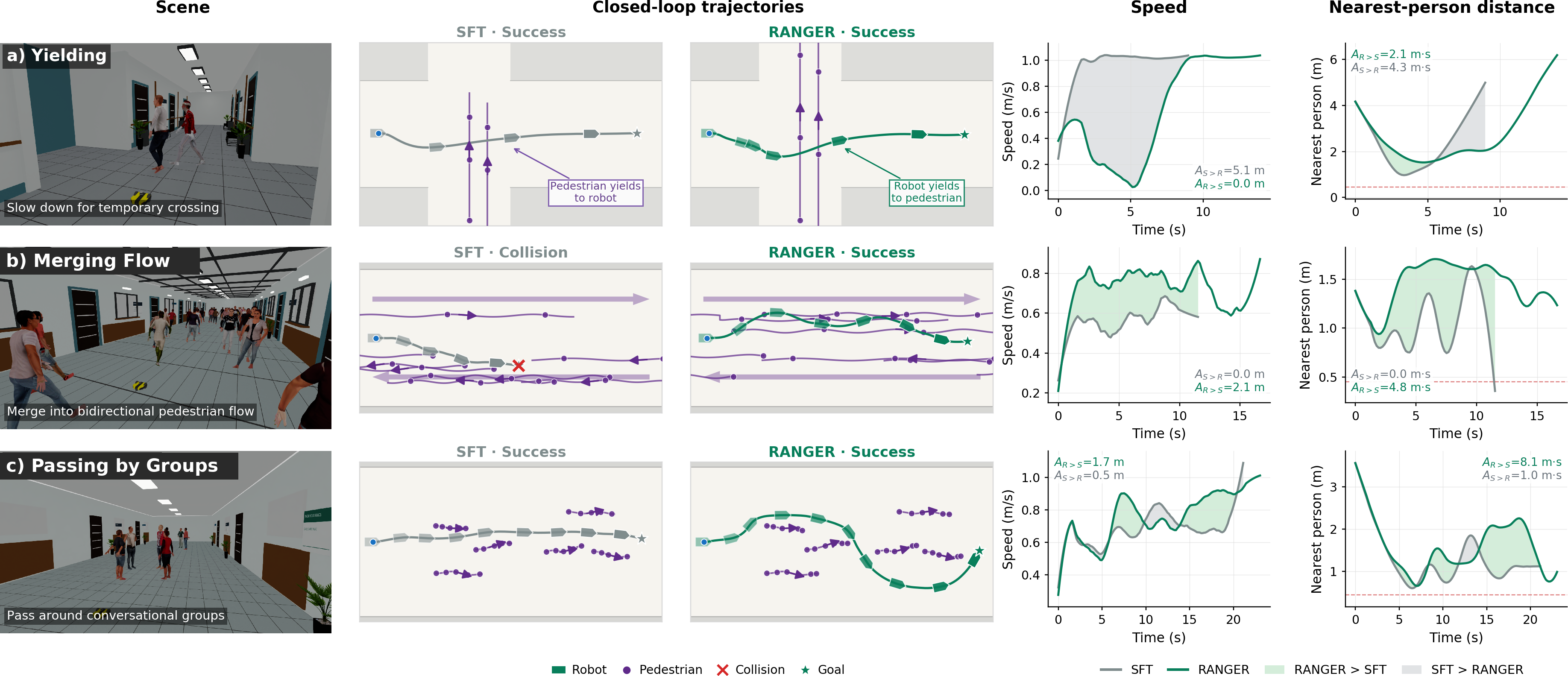}
    \caption{Detailed closed-loop diagnostics in three representative
    interaction scenarios: \emph{Yielding}, \emph{Merging Flow}, and
    \emph{Passing by Groups}.
    For each scenario, we show the scene snapshot, the closed-loop
    trajectories of the robot and pedestrians, and the corresponding
    speed and nearest-person-distance profiles for the SFT and
    post-trained RANGER policies.}
    \label{fig:task_grpo_qualitative_diagnostics}
\end{figure*}
\begin{table}[t]
\centering
\scriptsize
\setlength{\tabcolsep}{0.5pt}
\begin{tabular*}{\columnwidth}{@{\extracolsep{\fill}}lcccccc@{}}
\toprule
\multirow{2}{*}{Method}
& \multicolumn{1}{c}{\emph{Avoid.}}
& \multicolumn{2}{c}{\emph{Yielding}}
& \multicolumn{2}{c}{\emph{Merging}}
& \multicolumn{1}{c}{\emph{Groups}}\\
\cmidrule(lr){2-2}\cmidrule(lr){3-4}\cmidrule(lr){5-6}\cmidrule(lr){7-7}
& $d_{\mathrm{CA}}$ (m)$\uparrow$
& $\Delta v_{\mathrm Y}$ (m/s)$\uparrow$
& $d_{\mathrm Y}$ (m)$\uparrow$
& $e_{\mathrm{MF}}^{p}$ (m)$\downarrow$
& $e_{\mathrm{MF}}^{v}$ (m/s)$\downarrow$
& $d_{\mathrm{PG}}$ (m)$\uparrow$\\
\midrule
X-NavDP & 0.596 & 0.229 & \textbf{1.568} & 1.833 & \textbf{0.204} & 1.988\\
Flow-GRPO & \underline{0.910} & \underline{0.461} & 1.130 & \underline{1.071} & \underline{0.238} & \underline{2.279}\\
RW & 0.563 & 0.112 & 0.455 & 1.977 & 0.298 & 1.497\\
\textbf{RANGER} & \textbf{0.929} & \textbf{0.467} & \underline{1.395} & \textbf{0.911} & 0.269 & \textbf{2.572}\\
\bottomrule
\end{tabular*}
\caption{Skill-specific closed-loop performance across
four social navigation skills, evaluated using
pedestrian clearance, yielding, flow alignment,
and group-passing metrics.}
\label{tab:skill_closedloop_metrics}
\end{table}

\noindent\phantomsection\label{app:contents_qualitative}%
\textbf{Qualitative Analysis of Learned Behaviors.}
To examine how post-training changes navigation behaviors,
we compare closed-loop rollouts of the SFT and RANGER
policies in three representative interaction scenarios:
\emph{Yielding}, \emph{Merging Flow}, and
\emph{Passing by Groups}.
Fig.~\ref{fig:task_grpo_qualitative_diagnostics}
shows their executed trajectories, speed profiles,
and nearest-person distances under matched settings.
In \emph{Yielding}, both policies succeed but exhibit
different interaction behaviors. SFT proceeds while
the pedestrian yields, whereas RANGER slows down to
let the pedestrian cross before resuming motion.
The speed and distance profiles illustrate this yielding
behavior and a larger minimum pedestrian distance.
In \emph{Merging Flow}, SFT collides with a pedestrian,
whereas RANGER successfully merges into the surrounding
flow. Its trajectory aligns with pedestrian motion,
maintaining forward progress and safe separation
throughout the interaction.
In \emph{Passing by Groups}, both policies succeed,
but RANGER takes a wider detour around the
conversational group, maintaining greater clearance
from the occupied social space. Its smooth speed
profile further indicates that the interaction is
resolved primarily through spatial adjustment
rather than unnecessary stopping.
These results demonstrate that post-training enables
context-dependent adaptations in both trajectory
geometry and motion timing, allowing a single
RANGER policy to exhibit distinct social navigation
behaviors.

\begin{figure*}[t]
    \centering
    \begin{subfigure}[t]{\textwidth}
        \centering
        \includegraphics[width=\linewidth]{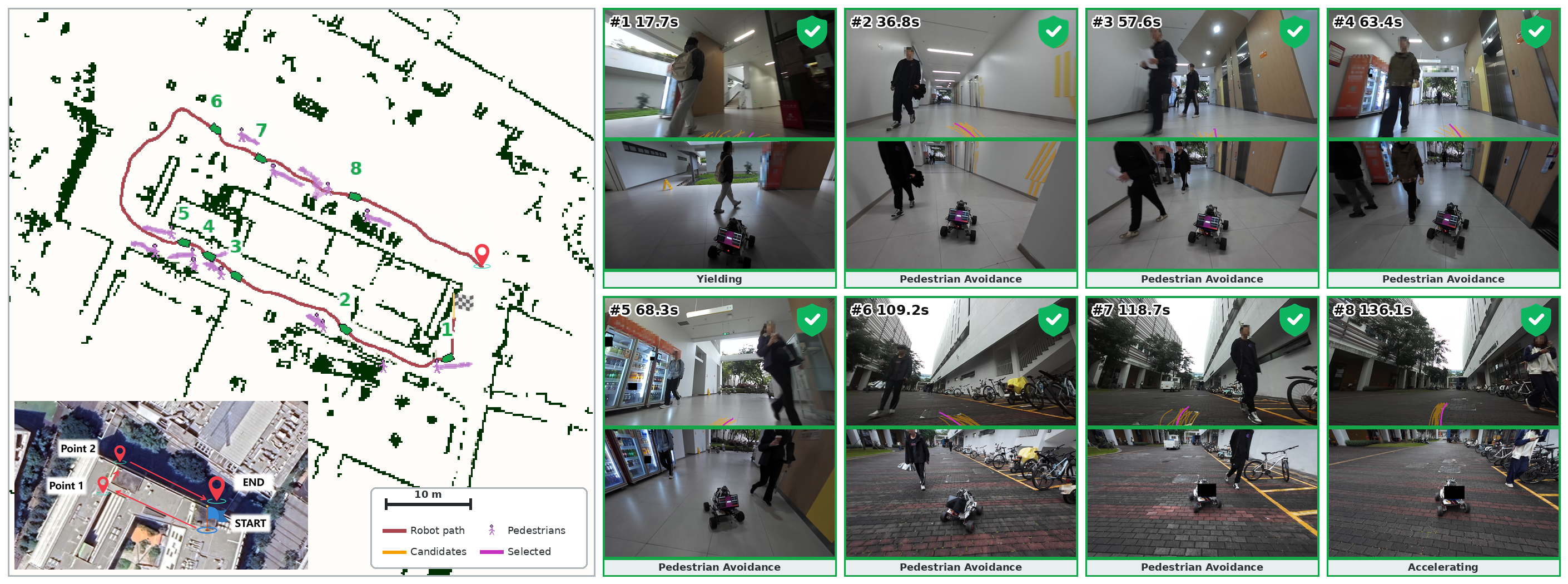}
        \caption{RANGER.}
    \end{subfigure}

    \vspace{0.8em}
    \begin{subfigure}[t]{\textwidth}
        \centering
        \includegraphics[width=\linewidth]{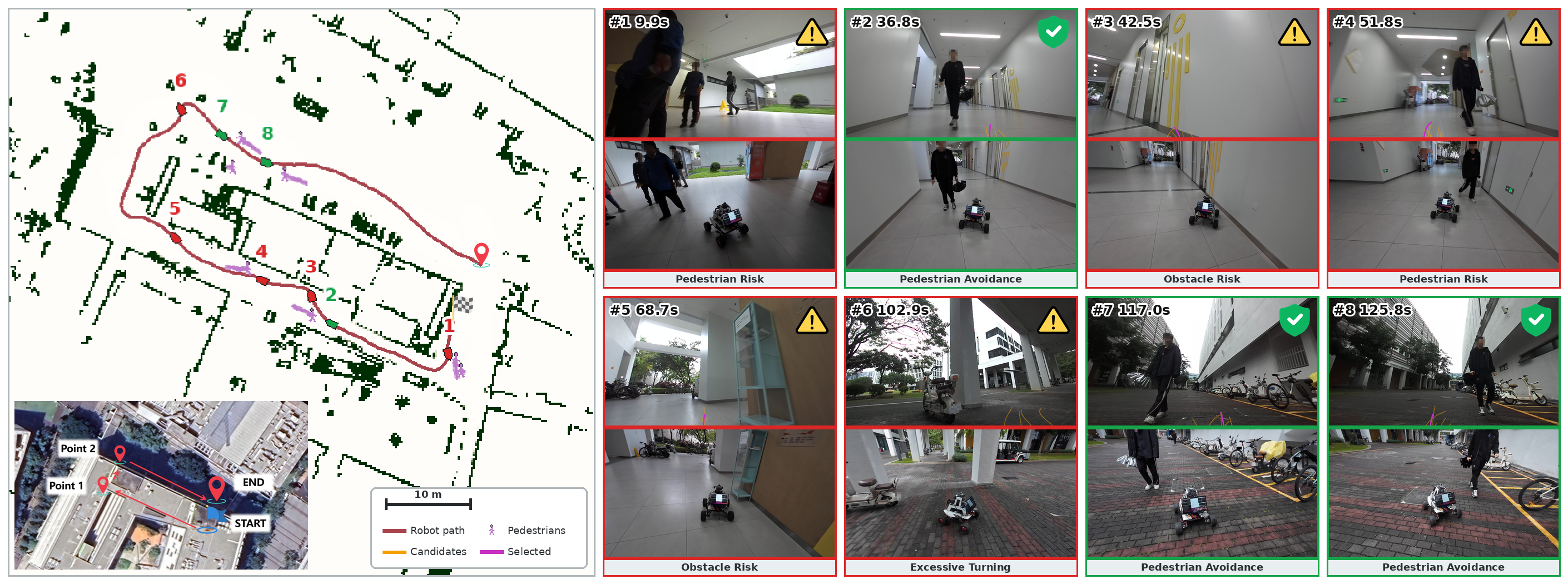}
        \caption{X-NavDP.}
    \end{subfigure}
    \caption{Real-world patrol deployment comparison between
    X-NavDP and RANGER.
    In each row, the left panel shows the complete executed
    robot path, pedestrian motion histories, and an aerial
    inset of the patrol route with two intermediate waypoints.
    The right panels present eight numbered locations along
    the route, each with a first-person observation (top)
    and a corresponding third-person view (bottom).
    Candidate and selected trajectories are overlaid on
    the observations. Green and red borders mark qualitatively
    safe and potentially unsafe robot behaviors, respectively,
    based on the selected trajectories and observed motion.}
    \label{fig:classroom_deployment}
\end{figure*}

\section{Additional Real-World Deployment}
\label{app:classroom_deployment}

\noindent\phantomsection\label{app:contents_deployment_setup}%
\textbf{Deployment Setup.}
To evaluate the reliability and adaptability of
post-trained navigation policies during extended
real-world deployment, we compare RANGER and
X-NavDP in a long-horizon autonomous patrol task.
The approximately $110\,\mathrm{m}$ route spans indoor
corridors, building entrances, and outdoor pedestrian
areas with randomly moving pedestrians.
Two intermediate waypoints are sequentially provided
as point goals before the final destination.
Both post-trained policies receive only onboard
RGB--D observations and the current point goal,
without access to a global map or privileged
pedestrian information.
The map in Fig.~\ref{fig:classroom_deployment}
is used solely for trajectory visualization.
Both methods complete the entire patrol autonomously,
without human intervention or manually specified
interaction behaviors.

\noindent\phantomsection\label{app:contents_behavioral_analysis}%
\textbf{Behavioral Analysis.}
Fig.~\ref{fig:classroom_deployment} presents
eight representative decisions from each deployment,
revealing distinct navigation behaviors along the route.
At the initial pedestrian encounter (\#1), RANGER
slows down and yields to a crossing pedestrian,
whereas X-NavDP exhibits pedestrian collision risk.
In the narrow corridor (\#2), both methods
successfully avoid an approaching pedestrian.
However, in the next encounter (\#3), X-NavDP generates a trajectory
with obstacle collision risk near the corridor wall,
while RANGER safely navigates around nearby
pedestrians. In the following encounter (\#4), X-NavDP again exhibits
pedestrian collision risk, whereas RANGER
maintains safe pedestrian avoidance.
In these representative encounters, X-NavDP generates
several risk-associated trajectories in confined spaces,
whereas RANGER exhibits more consistent avoidance behaviors.

Similar differences emerge during the transition
from indoor to outdoor environments.
Near the building entrance (\#5), X-NavDP
generates a trajectory with obstacle collision
risk, whereas RANGER adjusts its motion to
avoid a nearby pedestrian.
Near the outdoor passage (\#6), X-NavDP exhibits excessive turning,
indicating less stable
motion generation, while RANGER continues
navigating around pedestrians along the walkway.
In the outdoor pedestrian area (\#7), both
methods successfully avoid nearby pedestrians.
After the final pedestrian encounter (\#8), RANGER accelerates
as the path becomes clear, whereas
X-NavDP continues to exhibit avoidance behavior.
This highlights RANGER's ability to adapt both
its spatial trajectory and motion timing to
changing interaction conditions.

Overall, this long-horizon deployment highlights
RANGER's ability to adapt across indoor and outdoor
environments and autonomously transition between
yielding, pedestrian avoidance, and acceleration.
Compared with X-NavDP, which exhibits repeated
pedestrian and obstacle collision risks and
excessive turning, RANGER generates more consistent
and context-appropriate navigation behaviors.
These behaviors emerge from a single visual policy
without explicit skill labels or manual behavior
switching, supporting its practical adaptability
in extended real-world autonomous navigation.

\section{DynBench Closed-Loop Evaluation}
\label{app:dynbench_qualitative}

\noindent\phantomsection\label{app:contents_scene_design}%
\textbf{Scene and Pedestrian Design.}
We build our evaluation on DynBench~\cite{gong2026flux},
retaining its Isaac Sim environments, robot embodiment,
pedestrian assets, and motion simulation framework.
Importantly, the evaluation environments, pedestrian
appearances, and motion generation mechanisms are
distinct from those used in our SFT and post-training
data generation, providing independent visual and
dynamic conditions for evaluating generalization.

Full Warehouse retains 100 original DynBench episode scripts.
To reduce the influence of potentially infeasible routes, we
generate 100 episodes each for Hospital and Office under the
stated clearance constraints.
Robot start--goal routes range from $8$ to $22\,\mathrm{m}$,
with a minimum map clearance of $0.50\,\mathrm{m}$.
Pedestrian routes satisfy the same clearance, and their
initial positions maintain at least $1.2\,\mathrm{m}$
separation from one another and from robot endpoints.
To encourage frequent interactions, three pedestrian
routes are initialized within $3\,\mathrm{m}$ of the
robot route. Pedestrians repeatedly traverse their
routes in both directions.


\noindent\phantomsection\label{app:contents_robot_protocol}%
\textbf{Robot Deployment and Evaluation Protocol.}
All baseline methods are deployed using their respective
official code repositories and released model checkpoints.
We perform method-specific integration, parameter tuning,
and controller adaptation to accommodate the benchmark
robot and simulation environment, while preserving each
method's original policy architecture and inference
mechanism. These adaptations aim to ensure that each
method operates effectively under the shared benchmark
conditions rather than being disadvantaged by implementation
or embodiment mismatches.

All methods use matched evaluation episodes, sensor
configurations, robot dynamics, and speed limits, while
retaining their native trajectory generation and selection
strategies. The RGB--D camera is mounted at
$0.60\,\mathrm{m}$ with zero pitch and roll, a
$69^\circ$ horizontal field of view, and a resolution
of $640\times360$.
An episode terminates upon reaching the goal within
$1.5\,\mathrm{m}$, a pedestrian collision at a planar
distance below $0.45\,\mathrm{m}$, or the simulation
time limit.

\begin{figure*}[t]
    \centering
    \includegraphics[width=\textwidth]{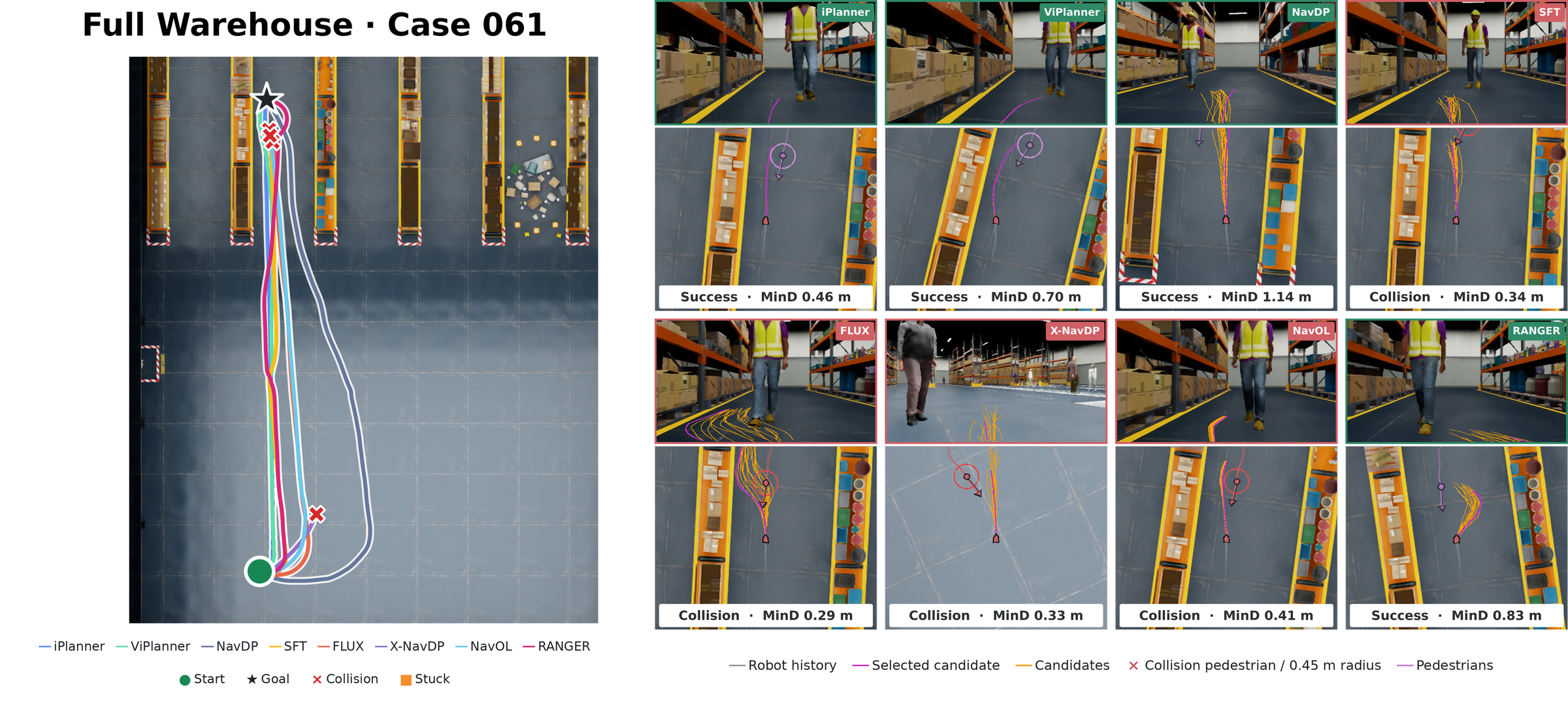}
    \caption{Qualitative comparison in Full Warehouse (Case 061).
    RANGER makes a smooth lateral adjustment to avoid the oncoming
    pedestrian and successfully reaches the goal with a minimum
    clearance of 0.83 m. While iPlanner, ViPlanner, and NavDP also
    succeed, SFT and the other post-trained baselines collide
    with the pedestrian.}
    \label{fig:dynbench_warehouse61}
\end{figure*}

\begin{figure*}[t]
    \centering
    \includegraphics[width=\textwidth]{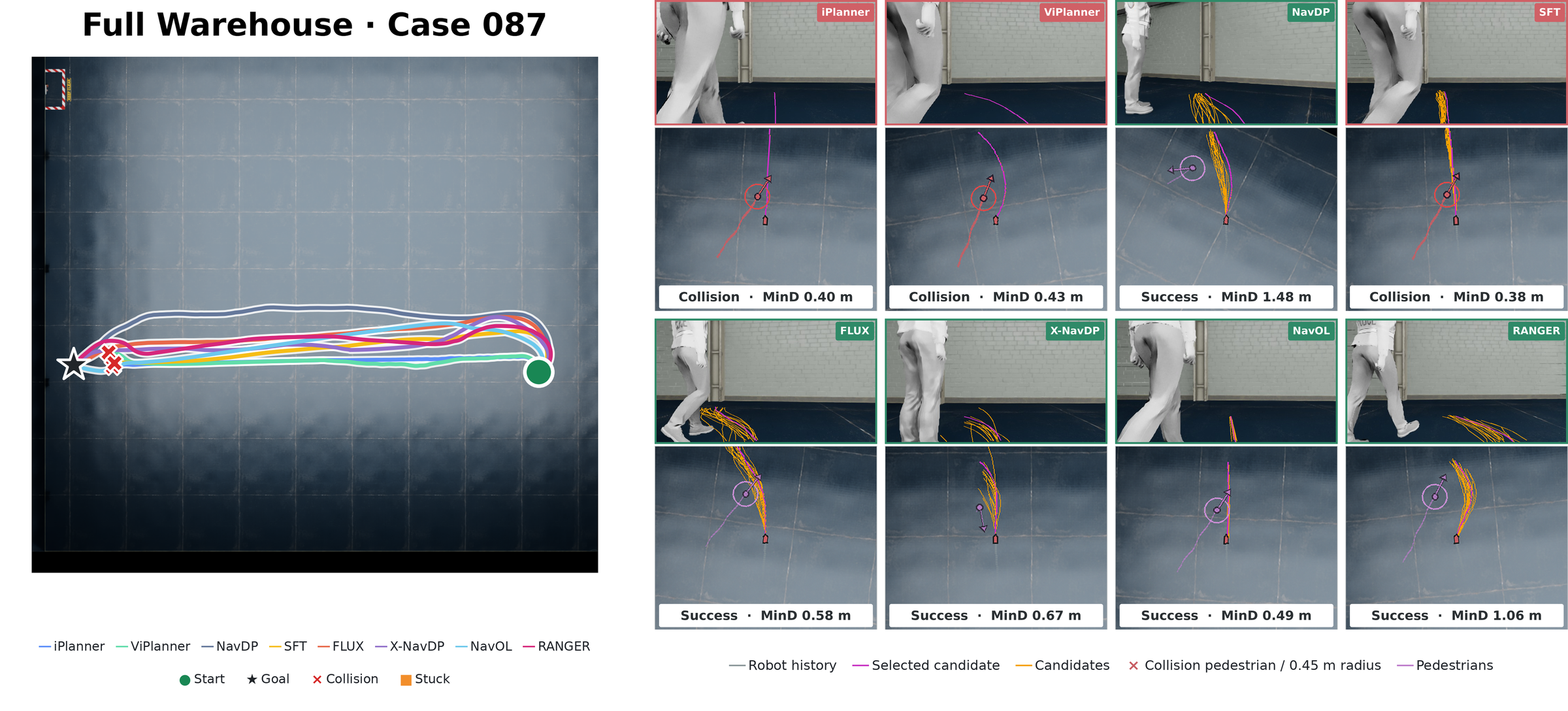}
    \caption{Qualitative comparison in Full Warehouse (Case 087).
    RANGER produces a smooth avoidance maneuver and reaches the goal
    with a minimum pedestrian clearance of $1.06\,\mathrm{m}$.
    In contrast, iPlanner, ViPlanner, and SFT collide during the
    close-range interaction.}
    \label{fig:dynbench_warehouse87}
\end{figure*}

\begin{figure*}[t]
    \centering
    \includegraphics[width=\textwidth]{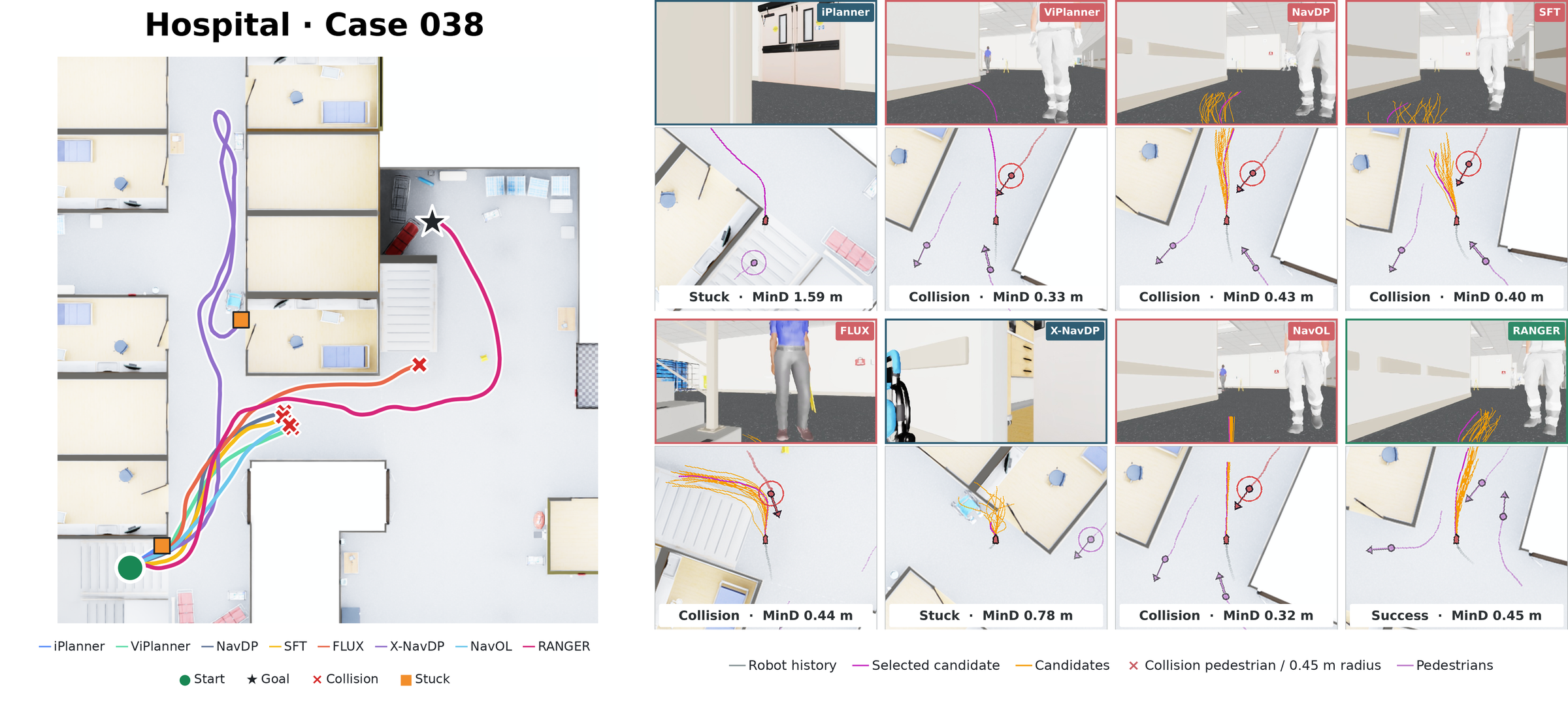}
    \caption{Qualitative comparison in Hospital (Case 038).
    RANGER takes a wider detour around the pedestrian interaction
    and successfully reaches the goal. In contrast, most competing
    methods collide near the corridor intersection, while iPlanner
    and X-NavDP become stuck.}
    \label{fig:dynbench_hospital38}
\end{figure*}

\begin{figure*}[t]
    \centering
    \includegraphics[width=\textwidth]{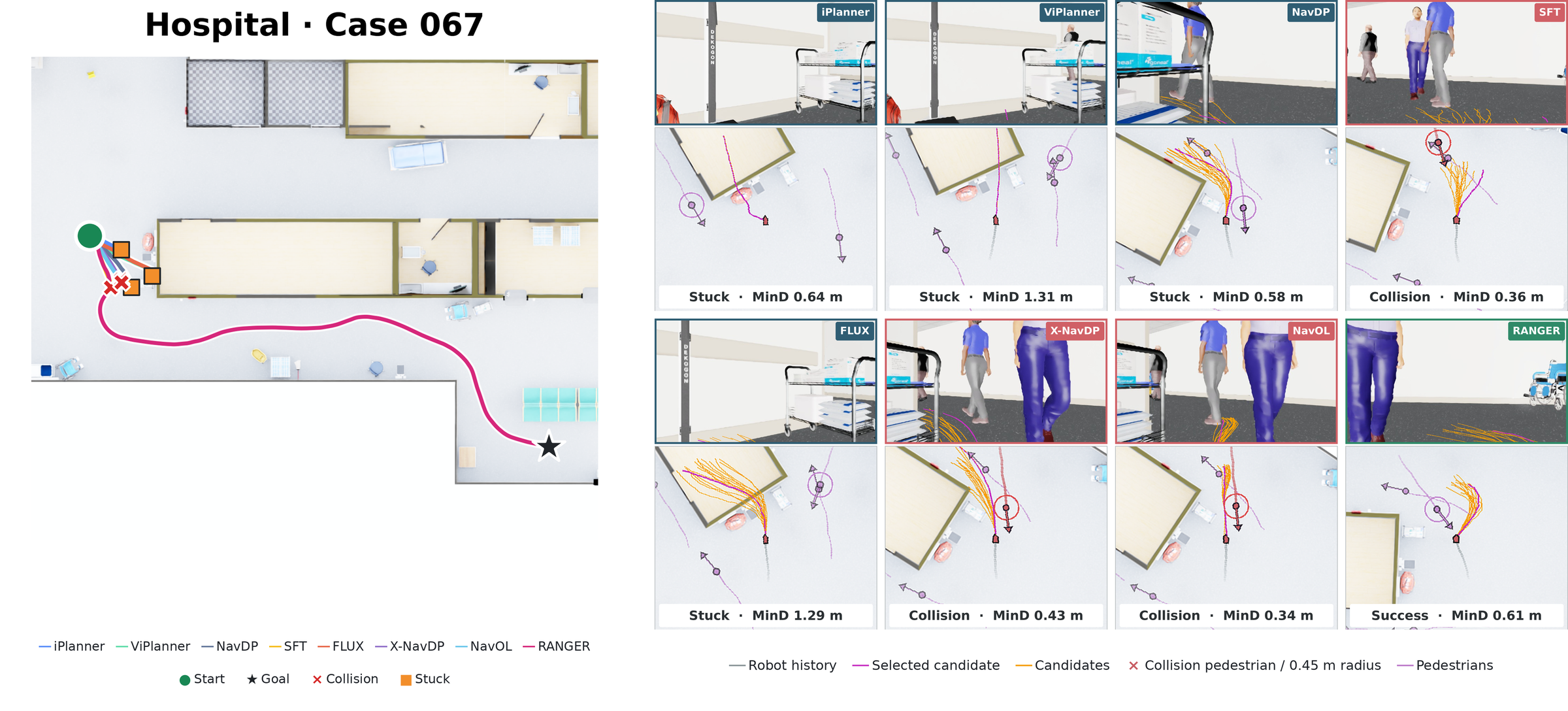}
    \caption{Qualitative comparison in Hospital (Case 067).
    RANGER selects an alternative route around the central
    obstacle and successfully navigates toward the goal while
    maintaining pedestrian clearance. All seven baselines fail,
    either becoming stuck near the initial passage or colliding
    with pedestrians.}
    \label{fig:dynbench_hospital67}
\end{figure*}

\begin{figure*}[t]
    \centering
    \includegraphics[width=\textwidth]{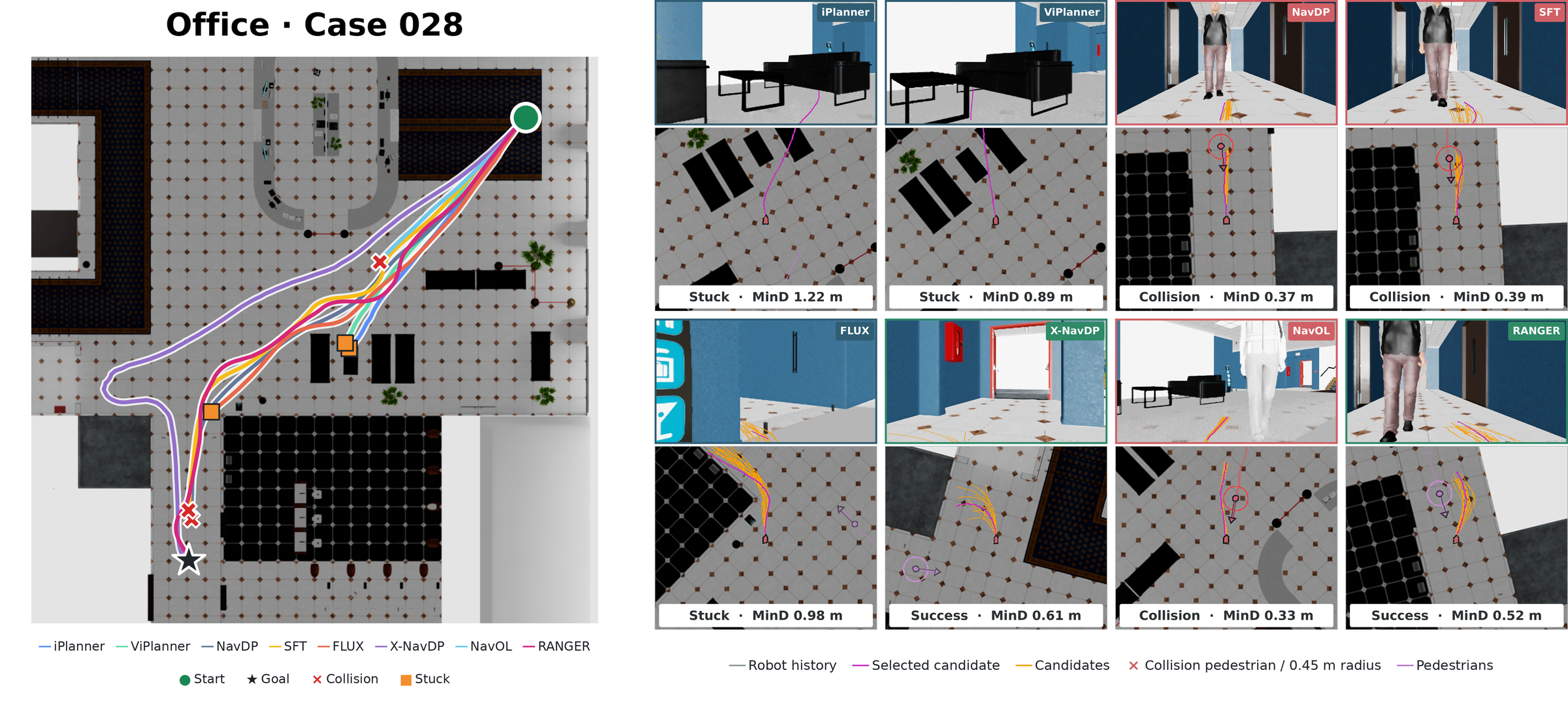}
    \caption{Qualitative comparison in Office (Case 028).
    RANGER follows a relatively direct route through the office
    while avoiding pedestrian collisions, successfully reaching
    the goal. X-NavDP also succeeds but takes a substantially
    longer detour, whereas the remaining methods either collide
    with pedestrians or become stuck.}
    \label{fig:dynbench_office28}
\end{figure*}

\begin{figure*}[t]
    \centering
    \includegraphics[width=\textwidth]{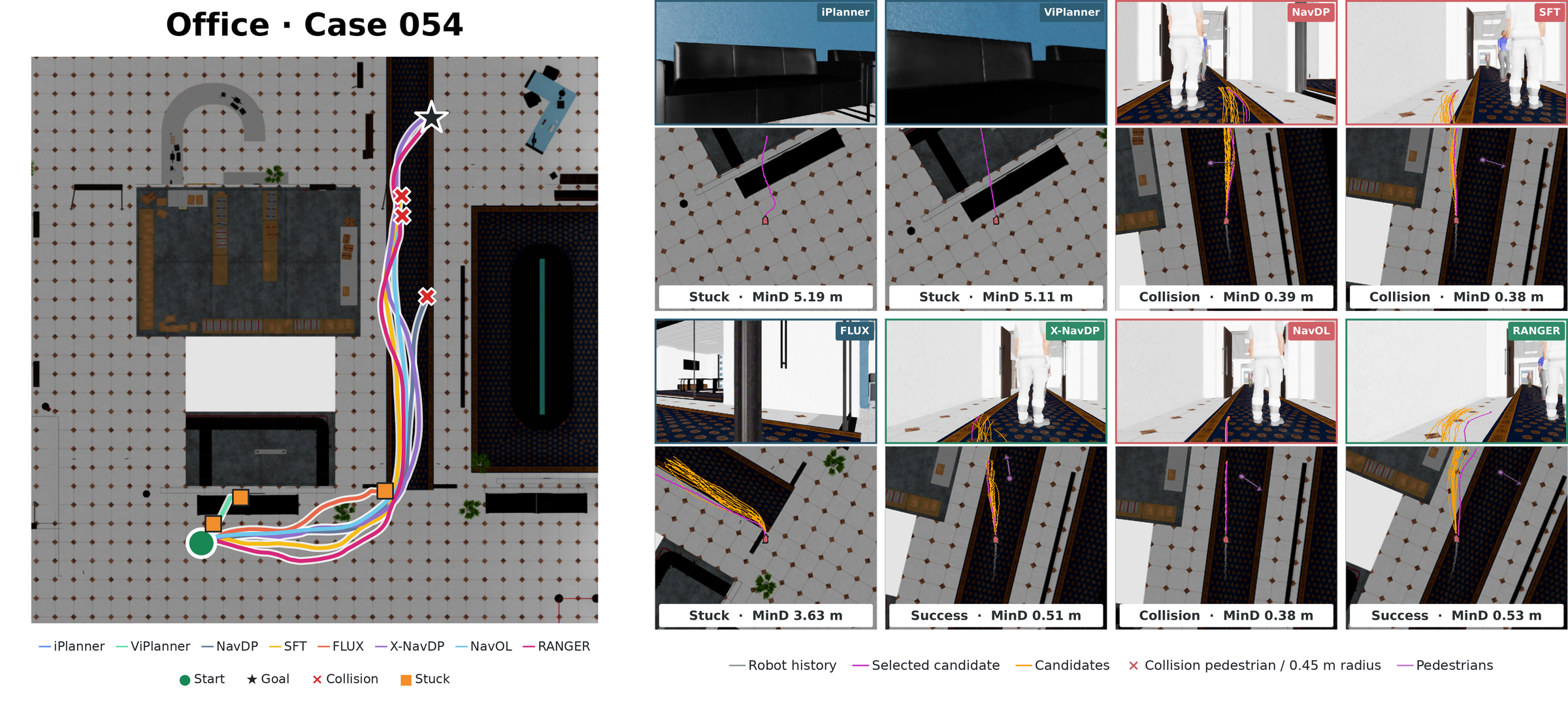}
    \caption{Qualitative comparison in Office (Case 054).
    RANGER navigates through the narrow corridor while maintaining
    a safe separation from pedestrians, reaching the goal with
    a minimum clearance of 0.53 m. X-NavDP also succeeds, whereas
    the remaining baselines either collide or fail to make
    sufficient progress.}
    \label{fig:dynbench_office54}
\end{figure*}

\noindent\phantomsection\label{app:contents_dynbench_comparisons}%
\textbf{Qualitative Comparisons.}
To further examine navigation behaviors and failure modes
across different methods,
Figs.~\ref{fig:dynbench_warehouse61}--\ref{fig:dynbench_office54}
present closed-loop comparisons on two Full Warehouse,
two Hospital, and two Office episodes. For each method, the first-person
observation is paired with a robot-centric top-down view
showing the executed history, sampled candidates, selected
trajectory, and nearby pedestrian motion. The local maps
are oriented with the robot heading upward for consistent
comparison. The left panel of each figure overlays the
complete robot trajectories of all eight methods on the
global map, with shared start and goal positions.

RANGER successfully completes all six episodes, while
other methods frequently collide with pedestrians or
become stuck. In Full Warehouse Case 61, RANGER reaches the goal while
SFT, FLUX, X-NavDP, and NavOL collide. In Case 87, iPlanner,
ViPlanner, and SFT collide, whereas RANGER reaches the goal with
$1.06\,\mathrm{m}$ minimum clearance. In the Office scenarios, both RANGER and
X-NavDP reach the goal, whereas RANGER is the only
successful method in the two Hospital scenarios.
The local trajectory visualizations further show that
RANGER maintains progress while avoiding nearby pedestrians.
In contrast, some baselines generate collision-prone
trajectories, while others remain stuck despite maintaining
large pedestrian clearance. These comparisons highlight
RANGER's ability to balance collision avoidance and
navigation progress in challenging dynamic interactions.

\section{Limitations}
\label{app:limitations}

Due to limitations of the current simulator, pedestrian behaviors during online post-training do not fully account for their interactions with the robot. In real-world environments, humans and robots continuously influence each other's motion and navigation decisions, leading to complex interaction dynamics such as mutual yielding, negotiation of shared space, and adaptation to changes in others' intentions. The limited modeling of such reciprocal interactions may restrict the acquisition of more sophisticated social navigation behaviors. Moreover, our post-training framework relies on privileged simulator information and manually designed skill-specific objectives, rather than autonomously discovering social norms from human behavior. Future work will investigate more realistic interactive simulation environments and autonomous learning and distillation of social navigation norms, enabling more adaptive navigation across diverse and previously unseen human--robot interactions.

\end{document}